\newcommand*{\ARXIV}{} 

\ifdefined\ARXIV
    \documentclass{article}
    \usepackage[utf8]{inputenc}
    \usepackage{arxiv}
    \usepackage{moreverb,url}
    \usepackage{amssymb}
\fi
\usepackage[english]{babel}
\usepackage{multirow}
\usepackage{xargs} 
\usepackage{comment}
\usepackage{amsmath,amssymb}
\usepackage{upquote}
\usepackage{algorithm}
\usepackage{csquotes}
\usepackage{ctable}
\usepackage{booktabs}
\usepackage{subcaption,graphicx}
\usepackage[flushleft]{threeparttable}
\usepackage[style=nature, backend=bibtex, natbib=true, hyperref = true, url=false, isbn=false, doi=false]{biblatex}
\usepackage{graphicx}
\ifdefined\ARXIV
    \usepackage{hyperref}
    \hypersetup{colorlinks,allcolors=black}
\fi
\usepackage{algpseudocode}% \makeatletter
\makeatletter
\algnewcommand{\LineComment}[1]{\Statex \hskip\ALG@thistlm \(\triangleright\) #1}
\makeatother
\newcommand{\secref}[1]{Sec.~\ref{#1}}
\newcommand{\tabref}[1]{Tab.~\ref{#1}}
\newcommand{\figref}[1]{Fig.~\ref{#1}}
\newcommand*{\algoref}[1]{Alg.~\ref{#1}}
\renewcommand{\eqref}[1]{Eq.~\ref{#1}}

\def\BibTeX{{\rm B\kern-.05em{\sc i\kern-.025em b}\kern-.08em
    T\kern-.1667em\lower.7ex\hbox{E}\kern-.125emX}}

\newcommand{\titleName}{Controlled  Generation of Unseen Faults for \textit{Partial} and \textit{Open-Partial} Domain Adaptation}
\newcommand{\abstr}{% THE PROBLEM
New operating conditions can result in a significant performance drop of fault diagnostics models due to the domain shift between the training and the testing data distributions. While several domain adaptation approaches have been proposed to overcome such domain shifts, their application is limited if the fault classes represented in the two domains are not the same. To enable a better transferability of the trained models between two different domains, particularly in setups where only the healthy data class is shared between the two domains, we propose a new framework for \textit{Partial} and \textit{Open-Partial} domain adaptation based on generating distinct fault signatures with a Wasserstein GAN. The main contribution of the proposed framework is the controlled synthetic fault data generation with two main distinct characteristics. Firstly, the proposed methodology enables to generate unobserved fault types in the target domain by having only access to the healthy samples in the target domain and faulty samples in the source domain. Secondly, the fault generation can be controlled to precisely generate distinct fault types and fault severity levels. 
The proposed method is especially suited in extreme domain adaption settings that are particularly relevant in the context of complex and safety-critical systems, where only one class is shared between the two domains. We evaluate the proposed framework on \textit{Partial} as well as \textit{Open-Partial} domain adaptation tasks on two bearing fault diagnostics case studies. Our experiments conducted in different label space settings showcase the versatility of the proposed framework. The proposed methodology provided superior results compared to other methods given large domain gaps }

\begin{document}

\ifdefined\ARXIV
    \title{\titleName}
    \author{%
        Katharina Rombach\\
        \textit{\small{Chair of Intelligent}} \\\textit{\small{Maintenance Systems}}\\
        ETH Z\"urich,\\
        Z\"urich, Switzerland\\
        \And 
        Gabriel Michau\\
        \textit{\small{Maintenance Systems}} \\\textit{\small{\& Technologies}}\\
        Stadler Service AG,\\
        Bussnang, Switzerland\\
        \And 
        Olga Fink\\
        \textit{\small{Intelligent Maintenance}} \\\textit{\small{and Operations Systems}}\\
        EPFL,\\
        Lausanne, Switzerland}
        \subtitle{Preprint}
        \date{\today}
        \maketitle
        \begin{abstract}
        \abstr
        \end{abstract}
\fi

\section{Introduction}
\label{sec:Introduction}
\begin{figure}[h]
\centering
\includegraphics[width=1.0
\textwidth]{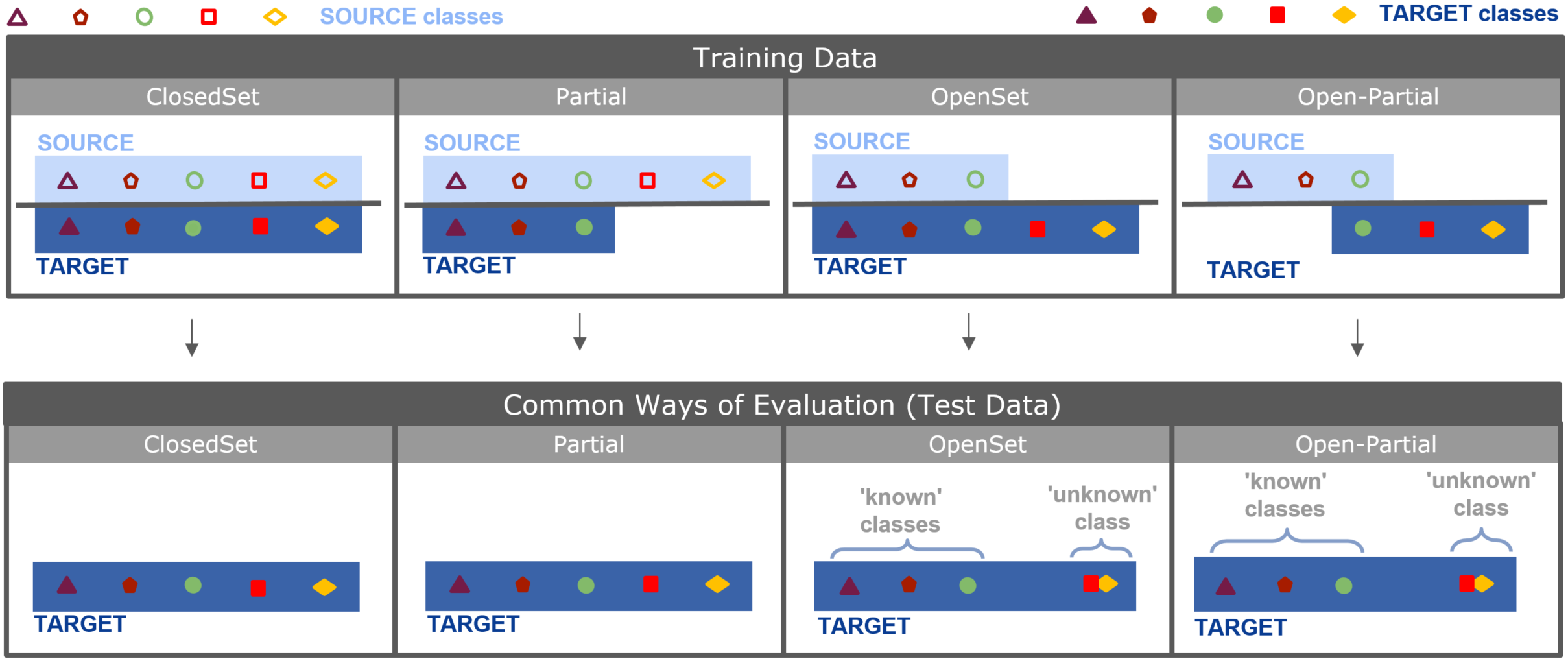}
\caption{\label{fig:Partial} Four DA configurations according to label
space discrepancies: (a) \textit{ClosedSet} ; (b) \textit{Partial}, (c) \textit{OpenSet},(d) both
\textit{Open-Partial} \cite{boris2021universal}.}
\end{figure}

A reliable operation of complex (safety-critical) assets can be achieved by monitoring the condition of the assets in real time, detecting the faults in an early stage and distinguishing between the different fault types to enable an informed schedule of the recovery maintenance or fault mitigation actions \cite{abid2021review}. % \cite{arunthavanathan2021analysis, liang2020novel}. 
Data-driven models based on real-time condition monitoring (CM) data  have shown a great potential for fault detection and diagnostics \cite{zhao2020deep, guan20212mnet}. However, CM data is often affected by distributional shifts (referred to as domain shifts), that can significantly decrease the performance of data-driven models \cite{miao2022sparse, zhou2022towards}. For example, changing operating conditions can cause such a distributional shift \cite{michau2021unsupervised, rombach2021contrastive}. 
Similarly, CM data of two units of a fleet can differ quite significantly due to differences in their configurations and operating regimes \cite{michau2019domain, li2021causal}.
To enable the transfer of a data-driven model to new operating conditions or new units in a fleet, domain adaptation (DA) methods have been successfully applied in fault diagnostics \cite{deng2022novel, lee2022asymmetric}. Most of  the proposed approaches, however, require that the same fault classes are represented in the source and the target domain. This DA setting, where the source and target domain datatsets cover the same classes, is referred to as \textit{ClosedSet} DA - see \figref{fig:Partial}. 
However, in real-world datasets, the classes represented in the two domains are not always congruent. 
 Due to the rareness of faults in complex industrial  assets, for example, observing each possible fault in all assets of a fleet and/or under all possible operating conditions 
 may not be practically feasible, particularly for safety-critical systems \cite{michau2021unsupervised}.
 Practical fault diagnostics solutions, typically, need to be taken into operation within a short period of time, not allowing to wait until all possible fault types have occurred. This results in cases where not all fault classes have been observed in all units or under all operating conditions, leading to label space discrepancies in CM datasets. 
 
\begin{figure}[h]
\centering
\subfloat[\centering Real source and target datasets in the extreme \textit{Partial} DA setting. \label{fig:challenge_1}]{\includegraphics[trim=10 0 600 0, clip, height=0.2\textwidth]{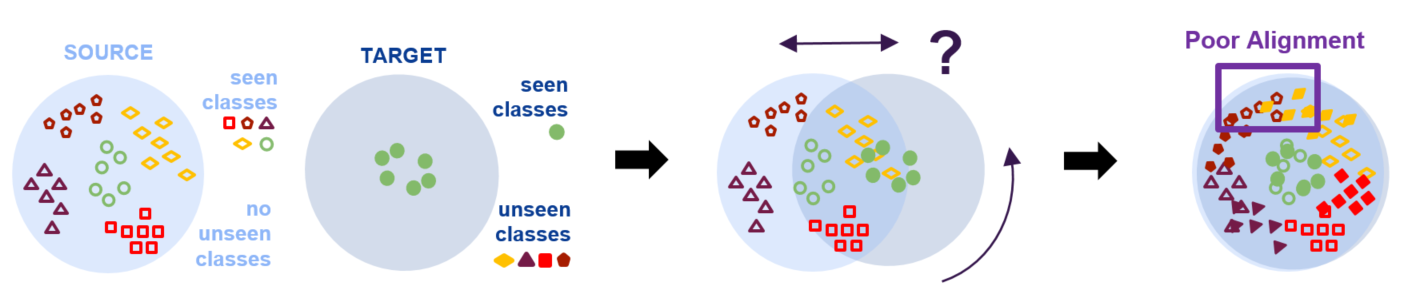}} 
{\includegraphics[trim=450 0 550 0, clip, height=0.2\textwidth]{images/CHALLENGE_ALIGN.png}} 
\subfloat[\centering Alignment problem. \label{fig:challenge_2}]{\includegraphics[trim=510 0 280 0, clip, height=0.2\textwidth]{images/CHALLENGE_ALIGN.png}}
{\includegraphics[trim=450 0 550 0, clip, height=0.2\textwidth]{images/CHALLENGE_ALIGN.png}} 
\subfloat[\centering A-posteriori evaluation. \label{fig:challenge_3}]{\includegraphics[trim=850 0 0 0, clip, height=0.2\textwidth]{images/CHALLENGE_ALIGN.png}}
\caption{\label{fig:challenge} Illustration of the source and target alignment challenge when only one class is shared between the domains on the example of the \textit{Partial} DA setting: The source and the target datasets are shown in \figref{fig:challenge_1} whereby only one class (green class) is represented in the target domain. The alignment step based on one class only is shown in \figref{fig:challenge_2}, whereby the challenge of finding the optimal alignment is indicated. The quality of chosen alignment method can only be tested during the a-posteriori evaluation, when the target classes have been observed ( see  \figref{fig:challenge_3}).}
\end{figure}

 In the literature, different DA settings have been distinguished by their type of  discrepancy in the label space \cite{boris2021universal}. This is illustrated in \figref{fig:Partial}. In the  \textit{Partial} DA setting, the target domain
covers only a subset of the source classes (source domain has private classes), whereas in the \textit{OpenSet} DA setting, the source domain covers only a subset of the target domain classes (target domain has private classes). The \textit{Open-Partial} DA setup is a combination of both previous settings where both domains have private classes that are not represented in the other domain.
Most existing DA methods are designed for only one of the above mentioned DA settings (mainly \textit{ClosedSet} DA) and are often not transferable to other DA scenarios
(see \secref{sec:RW}). 
This poses a challenge for successful DA in real applications, where different types of discrepancies in the label space can occur. 
Since safety-critical systems are reliable by design, faults occur very rarely. In some cases, only healthy data (one class) is available in different domains, not allowing to perform fault diagnostics at all. Instead, \citet{michau2021unsupervised}, proposed to train a one-class classification model that is transferable between domains. If faults did occur in one of the domains, fault diagnostic is possible. For the DA task, however, we are presented with an extreme case of label space discrepancy  in prognostics and health management (PHM) applications, where often only one class, the healthy one, is shared between the two domains \cite{wang2020missing}. 
For example, if a system starts operating under a new operating condition, only data of the assets's current condition will be available. For safety-critical systems, this is usually the healthy condition, meaning that only the healthy class is shared between datasets from various operating conditions (an extreme case of \textit{Partial}  DA). As illustrated in \figref{fig:challenge}, such an extreme case of label space discrepancy between two domains can pose a significant challenge for  \textit{Partial}  DA methods based on feature alignment. With only one class shared between the two domains, there exist many possible alignment solutions (see \figref{fig:challenge_2}) and their performance can only be evaluated after the model is employed and the real target faults have been observed (see \figref{fig:challenge_3}). 
Extreme discrepancies in the label space of training datasets can also arise if two units of a fleet 
are experiencing different fault types during the data collection (and model development) period. Then, in the available training dataset, the only common health class experienced so far by both units may be the healthy class. However, both of the units can be affected during their life times (during the deployment of the developed models) by the same failure modes. Therefore, the fault diagnostics algorithms should be able to diagnose all possible fault types and not only those that have been experienced by the specific unit at the training time. 
The results of previous studies show that, generally, the less classes are shared between the domains, the harder the DA task becomes \cite{wang2020missing, zhang2021open}.  For example, compared to the \textit{ClosedSet} setting, the classification performance on a bearing dataset dropped by 20\% when only three out of ten classes were shared between the two domains  \cite{zhang2021open}. 
% example here 
Despite the relevance to PHM applications, there is hardly any work tackling the extreme cases of discrepancies in the label spaces (with only one shared class between two domains) for fault diagnostics in different DA settings. These extreme scenarios are in the focus of the research in this paper. 

We propose to address the challenge raised by the label space discrepancies for DA by enabling the  generation of domain- and class-specific data from fault conditions that have not been observed before in the target domain. The generated fault data can compensate for unseen domain-specific fault classes and, thereby, transform the given \textit{Partial} or \textit{Open-Partial} DA setting into a \textit{ClosedSet} DA setting. 
The generation of previously unobserved target fault data is based on observed faults in the source domain i.e.\ we propose to perform unsupervised domain mapping.
 This is particularly challenging since it is unknown how an unobserved fault in the target domain should look like. The unsupervised target fault generation needs to fulfill two requirements.
 Firstly, the generated data should be adapted to the specificities of the desired domain and secondly, the faults should be specific to a desired class in the label space. We address for the first time such an unsupervised but controlled generation of fault data based only on the healthy data in the target domain and faulty data in the source domain. 
The proposed work is based on the hypothesis that the Fourier spectrum from  faulty data can be disentangled in data signatures that %are domain independent and 
represent (1)  solely fault class characteristics  and (2) domain-specific characteristics within the data. The validity of this hypothesis is evaluated implicitly by conducting different DA experiments. The main contribution of this research is a novel framework \textit{FaultSignatureGAN} based on a Wasserstein GAN \cite{arjovsky2017wasserstein} that enables to generate domain-independent fault class signatures that are  transferable to any new domain, given only healthy data of that domain. This is, on the one hand, a particularly challenging task since no samples of faulty data in the target domain  are available and, on the other hand, a particularly relevant case for real safety-critical applications where a representative dataset of fault data is typically not available.  \textit{FaultSignatureGAN} enables a controlled way to generate physically plausible faults of previously unobserved distinct classes in the target domain and thereby,
 enables to complement label spaces with different types of class discrepancies for DA tasks. Since the proposed framework relies solely on the availability of source faults and healthy target data, its benefits are particularly pronounced for targeting the extreme case of DA where only one class (the healthy class) is shared between the two domains. However, it is applicable to any number of shared and missing fault classes in the two domains.
The proposed framework \textit{FaultSignatureGAN} is not limited to only one type of label space discrepancy since it is applicable in \textit{Partial} as well as \textit{Open-Partial} setups. 

The remainder of the paper is organized as follows. First relevant related work is summarized in \secref{sec:RW}, the proposed framework is explained in \secref{sec:Methodology}. The case studies are introduced in \secref{sec:CaseStudies} and the exact setup of the conducted experiments is stated in \secref{sec:ExpDomain}. The results of the conducted experiments for \textit{Partial} DA are  shown in \secref{sec:Partial} and for \textit{Open-Partial} settings  in \secref{sec:OpenSet}. The findings are discussed in \secref{sec:discussion} and conclusions are drawn in \secref{sec:Conclusion}.

\section{Related Work}
\label{sec:RW}
\textbf{Domain Adaptation} has been intensively studied in recent years in the context of PHM applications \cite{li2022perspective}. Proposed approaches include discrepancy-based \cite{li2021central, zhang2020sparse} and
adversarial \cite{deng2022novel, wang2020triplet} methods.
%and reconstruction based methods \cite{wen2017new, sun2019sparse}.
Most of the proposed approaches, however, have been exclusively developed for the \textit{ClosedSet} DA setting where  the source and target domain cover the same classes - see left column in \figref{fig:Partial}.
As exemplified in \secref{sec:Introduction}, the assumption of a \textit{ClosedSet} setting is generally not realistic in many practical applications. This is particularly the case in safety-critical applications where faults are very rare \cite{michau2021unsupervised}. 
Hence, the previously mentioned \textit{ClosedSet} DA methods do not meet the requirements and conditions of industrial applications. There has been an increasing interest to develop methods that address more realistic DA scenarios with label space discrepancies to enable reliable fault diagnostics also for safety-critical applications.
% Computer Vision
Numerous approaches for \textit{Partial} \cite{liang2020balanced, cao2018partial}, \textit{OpenSet} \cite{li2021domain, pan2020exploring} as well as \textit{Open-Partial} \cite{saito2020universal, saito2021ovanet} DA approaches have been developed in the field of computer vision.
Recently, several research studies  have developed the ideas further to adapt them to the challenges of real CM data. 

 % OpenSet
In the context of fault diagnostics, adversarial approaches have been proposed for \textit{OpenSet} DA and experiments were conducted  with different degrees of label space discrepancies i.e.\ with a different number of shared classes between the two domains. For example,  \citet{zhao2022dual} introduced an auxiliary domain discriminator to attribute less weight to private target samples and emphasized the importance of developing methods that can deal with significant label space discrepancies. Experiments were conducted with as little as two shared classes between the two domains.  In the \textit{OpenSet} DA context, \citet{zhang2021open} used an instance-level weighted mechanism to identify private target classes and tested the proposed method i.a.\ with three (out of ten) shared classes between the domains. The results demonstrated that, generally, the less classes are shared between the domains, the harder the DA task becomes.  In another study on \textit{OpenSet} DA, extreme value theory was proposed to reject private target samples \cite{yu2021deep}. Another method proposed a source class-wise and target instance-wise weighting mechanism combined with an additional outlier
identifier. The proposed method has even been applied on multiple label space discrepancy settings \cite{zhang2021universal}. Despite having multiple classes shared between the two domains, the method's performance was negatively impacted when the target domain had private classes.
% Classification of unknown faults
One of the major limitations of the proposed approaches is that they only aim to classify known classes (i.e.\ source classes) and do not enable to distinguish private target samples into different classes. The same evaluation scheme, where private target samples are not distinguished by different classes but are considered as one unknown class,  is used in the field of computer vision \cite{saito2020universal}. In safety-critical applications, however, it is important to distinguish between different health conditions within the private target samples to plan appropriate maintenance actions. 
%Openess
Moreover, despite its relevance to fault diagnostics in  safety-critical complex technological system, none of the above mentioned  studies has tackled the extreme case of the \textit{OpenSet} DA setting, where only one class (the healthy class) is shared between the domains. 

% Partial
Methods targeting the \textit{Partial} DA setting have also been developed for PHM applications. For example, a class-weighted adversarial DA method was proposed that uses the domain discriminator's output to detect the private classes \cite{li2020partial}. The output of two classifiers has also been employed to estimate the target distribution and train domain-invariant representations \cite{jiao2019classifier}. Also, randomly selected source data is used to augment the target domain to align the conditional distributions combined with a class-wise adaptation \cite{zhao2022balanced}. 
% Openess
Some research studies have even dealt with the extreme case of \textit{Partial} DA where only the  healthy state is shared between the two domains \cite{li2020deep2, li2018cross, wang2020missing}. 
For example, \citet{li2020deep2} proposed a conditional data alignment step (using the maximum mean discrepancy) that is only applied to the healthy data from the source and target domain to prevent misalignment due to the label space discrepancy. Additionally to the conditional alignment, the authors proposed prediction consistency schemes using multiple classifier models for fault diagnostics in \textit{Partial} DA settings. \citet{wang2020missing} proposed an unilateral alignment approach (\textit{Unilateral})  for \textit{Partial} DA with extreme label space discrepancy. The proposed method made use of the inter-class relationships of the source domain and aligned the target features
to the pre-trained source domain features. Although the results of previous
studies using different feature source and target alignment techniques in \textit{Partial} DA settings with extreme label space discrepancies are promising, the methods have mainly been tested on CM datasets with small domain gaps (indicated by the high Baseline classification performance). The employed methods may fail under large domain shifts, where the inter-class relationships might have changed significantly. Another limiting factor in applying the above mentioned feature alignment methods to new safety-critical assets is finding an optimal hyperparameter setting. With only one class being shared between the domains, there exist multiple possible alignment solutions and their quality can only be evaluated a posterior, posing a safety risk in industrial assets. Therefore, previous works used, e.g., data and labels from target faults for one validation domain shift to tune the hyperparameter \cite{wang2020missing}. This solution to find the optimal hyperparameter setting is, however, not possible in real applications where data from unobserved target faults is not available.    
One fault data generation approach was investigated in the extreme case of \textit{Partial} DA combined with an additional alignment step \cite{li2018cross}. 
However, the proposed target data generation method required extrapolation abilities of the generative model. Given the limited extrapolation abilities of deep models, it is not to be expected that the generated data resembles realistic target faults - especially given large domain gaps. Instead of generating target data as performed in \cite{li2018cross}, \citet{zhao2022balanced} adapted the idea of \citet{liang2020balanced} and proposed to augment the target data with source data to compensate the missing class data and performed adversarial feature alignment on the augmented and class-weighted datasets combined with a class-center-alignment loss. While the source data augmentation stabilized the alignment process, the proposed method may fail in settings where the inter-class relationships might have changed significantly.
% Only applicable in one setting
The above mentioned approaches tackling different settings of label space discrepancy in DA have usually been developed for one specific DA setting, either\textit{Partial} or \textit{OpenSet}, and are typically not applicable in other settings. Furthermore, large domain gaps have not been tackled so far in the extreme case of label space discrepancy where only the healthy class is shared between the domains. 
 In this work, we aim to develop a framework that performs well in the extreme case of DA under different label space discrepancy settings with a particular focus on the \textit{Partial} and \textit{Open-Partial} setting. Further, we aim to enable DA also in the cases where the domain gaps are large.

 % Other approaches 
\textbf{Domain generalization} addresses the challenge of fault diagnostics under unforeseen domain shifts (contrary to one explicit shift between two domains). Different techniques have been proposed in the context of fault diagnostics \cite{zhou2022towards, zhao2022domain, li2021causal}, including an approach for the \textit{OpenSet} setting \cite{zhao2022adaptive}. However, these methods generally require access to multiple source domains, mainly with shared labeled spaces in the source domains. This is often not given in industrial applications and therefore, domain generalization approaches are not applicable to the challenges addressed in this research. 
 
\textbf{Controlled Synthetic Data Generation} has raised a lot of attention in recent years \cite{gui2021review}, especially in the field of computer vision. Generative models have also been applied for DA. In particular, conditional generative models have been employed for domain mapping i.e.\ to translate a source input image to an image that closely resembles the target distribution \cite{ wilson2020survey}. However, existing approaches require a \textit{ClosedSet} DA setting since the target domain typically inherits the labels from the source domain \cite{bousmalis2016domain}. 

In the context of PHM applications, generative models have mainly been  applied to balance imbalanced datasets \cite{mao2019imbalanced, zhang2020machinery}, whereby e.g.\  conditional GANs have been used  to control the generation process to generate desired distinct classes \cite{luo2021case}.  However, those approaches are solely suited to generate data from known classes: classes that have been observed before. The approaches are not suited to generate data from classes in a specific domain, where they have not been observed before. The latter is the focus of our research. 
\citet{wang2021integrating} addressed the challenge of fault diagnostics in safety-critical systems, where faults are extremely rare and only little fault data is available. Given a real unlabeled dataset with only a few real fault samples (highly imbalanced), a balanced synthetic fault dataset is generated based on expert knowledge about specific patterns of different fault types.  DA is then performed in a subsequent step  to close the synthetic-to-real domain gap. The expert knowlege, therefore, enables to transfer different fault types to different types of bearings. However, this approach requires  a substantial domain knowledge. Furthermore, patterns of different fault types (as in \cite{wang2021integrating}) are typically easier to distinguish compared to different severity levels of the same fault type, as addressed in this research. To distinguish between different fault severities as well as types, synthetic data representing also different fault severities is required.     

The concept of \textbf{disentenglement}, which is based on the hypothesis that real-world data is generated by a few independent explanatory factors of variation \cite{locatello2019challenging, hu2022remaining},   has also been studied in the context of controlled data generation \cite{deng2020disentangled}. Proposed methods, typically, aimed to learn a disentangled feature representation of the data and used these disentangled features representing independent factors of variation to generate data samples with desired characteristics in controlled ways  \cite{zhou2019talking, kelkar2022prior}. 
However, although disentangled representations should be general and are expected to be generalizable to new domains, recent studies  found that disentanglement does not guarantee combinatorial generalization (understand and produce novel combinations of familiar elements) \cite{liu2022learning, schott2021visual, montero2020role}. 
To mitigate the lack of generalizability of disentangled representations, it is possible to constrain the disantangelment using a priori knowledge on the data structure.
For example, \citet{yang2020fda} simply assumed that, in image datasets, different factors of variations can be distinguished by different ranges in the frequency spectrum: the domain-specific information is solely represented in the low frequency range whereas the semantic information is reflected only in the high frequency range. This assumption allowed \citet{yang2020fda} to generate unseen target data simply by swapping the domain-specific low-frequency block of the source and target images and perform DA with the synthetically generated data. 
This block-wise distinction into a domain-specific and a semantic-specific frequency ranges can be considered as a disentanglement in the Fourier space.  Unfortunately, such a block-wise distinction of fixed frequency ranges representing either solely the domain-specific or semantic-specific components is not possible for CM data from 
mechanical systems with complex dynamic behaviour.  Instead, we expect an fault as well as a domain shift to affect the entire frequency spectrum. 
Based on the intuition that the operating conditions are 
independent of defects, we can assume that faults create disturbances on top of existing signals. We, therefore, assume that the Fourier spectrum can be expressed as the sum of domain-specific components and fault-specific components. This assumption, that both information content (domain-specific and semantic-specific) does not only affect constrained frequency ranges but rather impact the entire spectrum, generalizes the work of \cite{yang2020fda} to data from other application domains, in particular to CM data from complex industrial systems. Further, the assumptions enables the generation of unseen data while neither relying on combinatorial generalization of disentangled features nor relying on extrapolation abilities of the  generative model. 

\textbf{In this work}, we aim to develop a framework that enables a controlled generation of novel distinct fault classes in a target domain where the fault condition has not been observed before. Thereby, the \textbf{contribution of our proposed framework} is the generation of unseen domain-specific fault data, that enables DA with extreme label space discrepancies, also under large domain gaps. Contrary to other generative approaches, we do not only control the class being generated but also the specific domain of the data. Further, the data generation is unsupervised since the respective target fault has not been observed before in the specific target domain i.e.\ we enable the controlled generation of out-of-distribution data. Although the developed framework enables the generation of previously unseen data, it does not rely on extrapolation abilities of the generative model  but instead, it relies on a disentanglement assumption. This assumption enables to transfer the fault information between different domains and, ultimately, to generate physically plausible data of unseen fault classes. The proposed framework, therefore, enables to generate data that can substitute for missing domain-specific class data for DA problems with label space discrepancies. The generated data will especially be useful to overcome the alignment challenge  in  cases of extreme label space discrepancies, e.g. with only one overlapping class between the two domains (see \secref{sec:Introduction}). It enables DA in multiple scenarios where the label space is initially not congruent. 
Our methodology is especially suited for DA in the extreme case of label space discrepancy
, where only one class is shared between the domains and thereby, addresses an important requirement of reliable fault diagnostics in complex industrial (safety-critical) assets. However, it is also applicable to DA setups with any number of missing classes. Furthermore, contrary to other DA approaches, the proposed methodology is universally (\textit{Partial}, \textit{Open-Partial}) applicable.

\section{Methodology}
\label{sec:Methodology}
\begin{figure}[h]
\centering
\subfloat[\centering Real data. \label{fig:partialsubs_1}]{\includegraphics[trim=0 0 850 0, clip, height=0.13\textwidth]{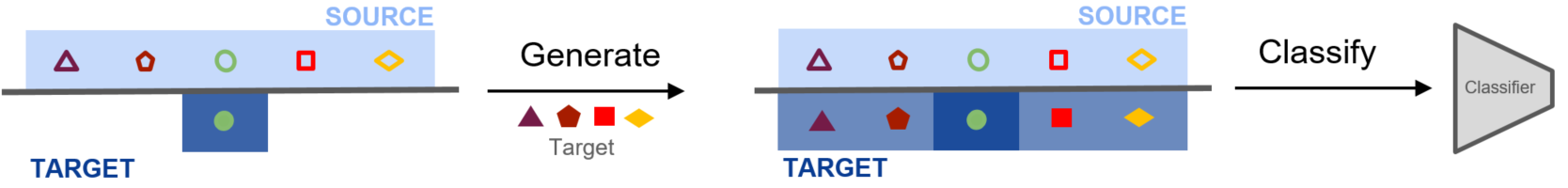}} 
\subfloat[\centering Generate data. \label{fig:partialsubs_2}]{\includegraphics[trim=400 0 680 0, clip, height=0.13\textwidth]{images/partial_meth.png}} 
\subfloat[\centering Real and generated data. \label{fig:partialsubs_3}]{\includegraphics[trim=600 0 280 0, clip, height=0.13\textwidth]{images/partial_meth.png}} 
\subfloat[\centering Classifier. \label{fig:partialsubs_4}]{\includegraphics[trim=980 0 0 0, clip, height=0.13\textwidth]{images/partial_meth.png}} 
\caption{\label{fig:partialsubs}  \textbf{FaultSignatureGAN} in the \textit{Partial} DA settings: the original data setting is depicted in \figref{fig:partialsubs_1}; the missing target classes are generated in \figref{fig:partialsubs_2}; the target dataset is augmented with synthetically generated data in \figref{fig:partialsubs_3} and a classifier is trained on the augmented dataset in \figref{fig:partialsubs_4}. }
\end{figure}

\begin{figure}[h]
\centering
\subfloat[\centering Real data. \label{fig:OSPAsubs_1}]{\includegraphics[trim=0 0 870 0, clip, height=0.13\textwidth]{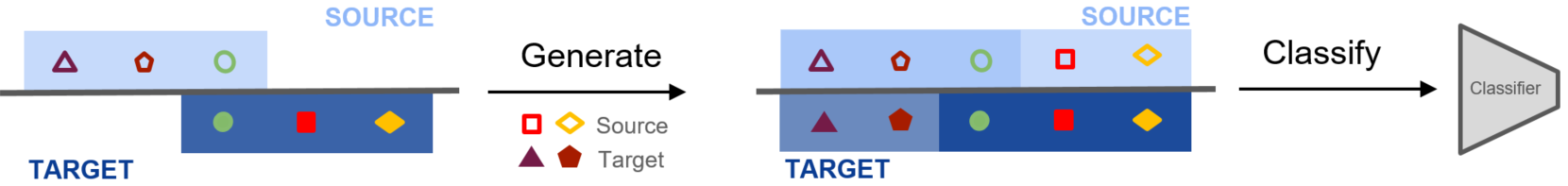}} 
\subfloat[\centering Generate data. \label{fig:OSPAsubs_2}]{\includegraphics[trim=380 0 680 0, clip, height=0.13\textwidth]{images/OSPA_meth.png}} 
\subfloat[\centering Real and generated data. \label{fig:OSPAsubs_3}]{\includegraphics[trim=600 0 280 0, clip, height=0.13\textwidth]{images/OSPA_meth.png}} 
\subfloat[\centering Classifier. \label{fig:OSPAsubs_4}]{\includegraphics[trim=960 0 0 0, clip, height=0.13\textwidth]{images/OSPA_meth.png}} 
\caption{\label{fig:OSPAsubs}  \textbf{FaultSignatureGAN} in the \textit{Open-Partial} DA settings: the original data setting is depicted in \figref{fig:OSPAsubs_1}; the missing source and target classes are generated in \figref{fig:OSPAsubs_2}; the source and target dataset is augmented with synthetically generated data in \figref{fig:OSPAsubs_3} and a classifier is trained on the augmented dataset in \figref{fig:OSPAsubs_4}.}
\end{figure}

We propose a framework, referred to as \textit{FaultSignatureGAN}, that enables to generate distinct domain-independent fault signatures based on the hypothesis that the faulty signal can be represented as the sum of domain-specific components and fault-specific components. These fault signatures can be transferred to new target domains such that the transferred data is  representative of distinct fault classes in a target domain where they have not been observed before. 
The generated data is then used in a subsequent step to substitute for missing class data in different DA settings with label space discrepancies: \textit{Partial} (see \figref{fig:partialsubs}) and \textit{Open-Partial} DA (see \figref{fig:OSPAsubs}). Finally, a classification model is trained on the augmented datasets.

\subsection{Training the generative model} 
\begin{figure}[h]
\centering
\makebox[\textwidth]{\includegraphics[trim=0 0 0 0, clip,height=0.32\textwidth]{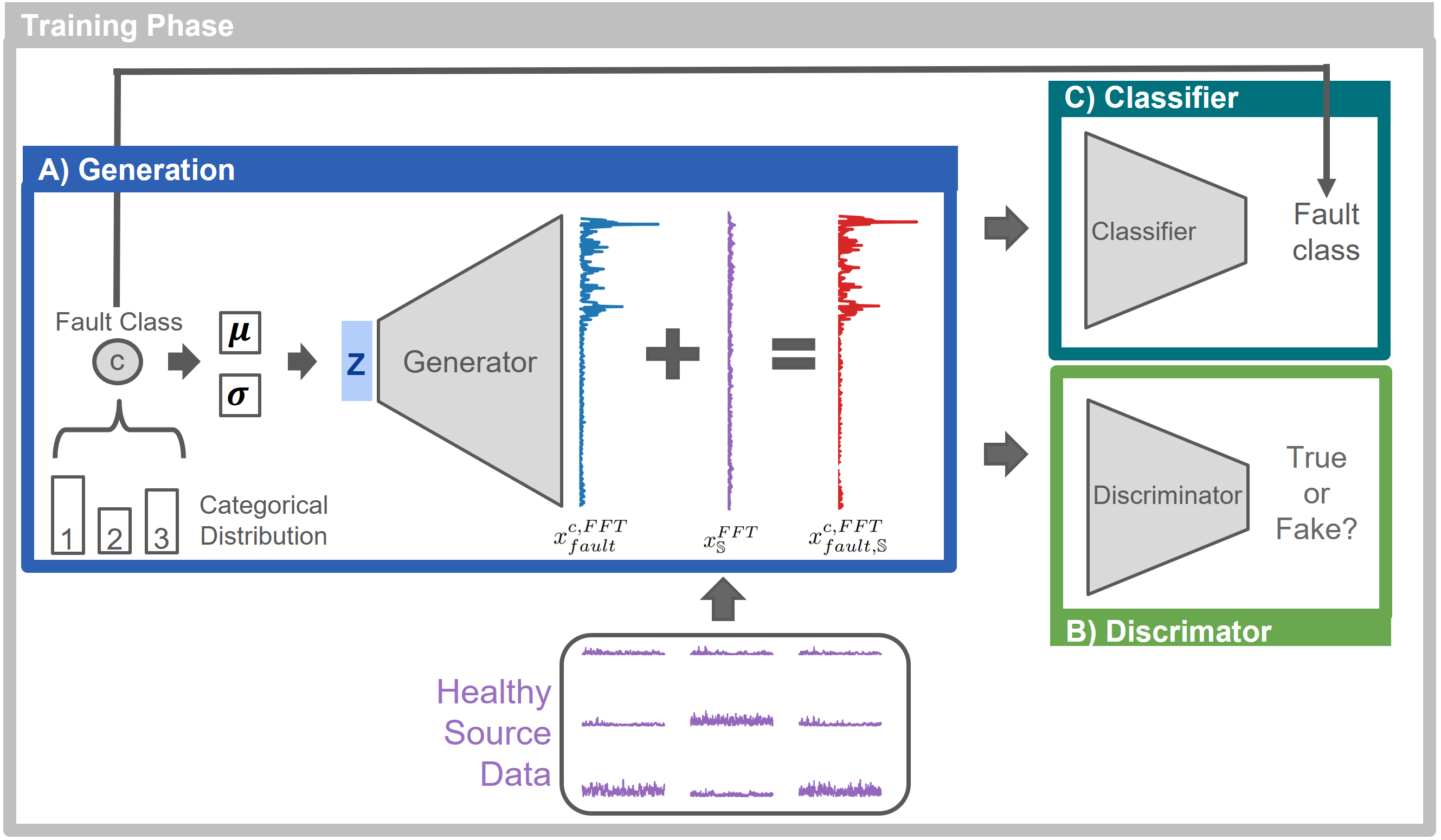}
\includegraphics[trim=0 0 0 0, clip, height=0.32\textwidth]{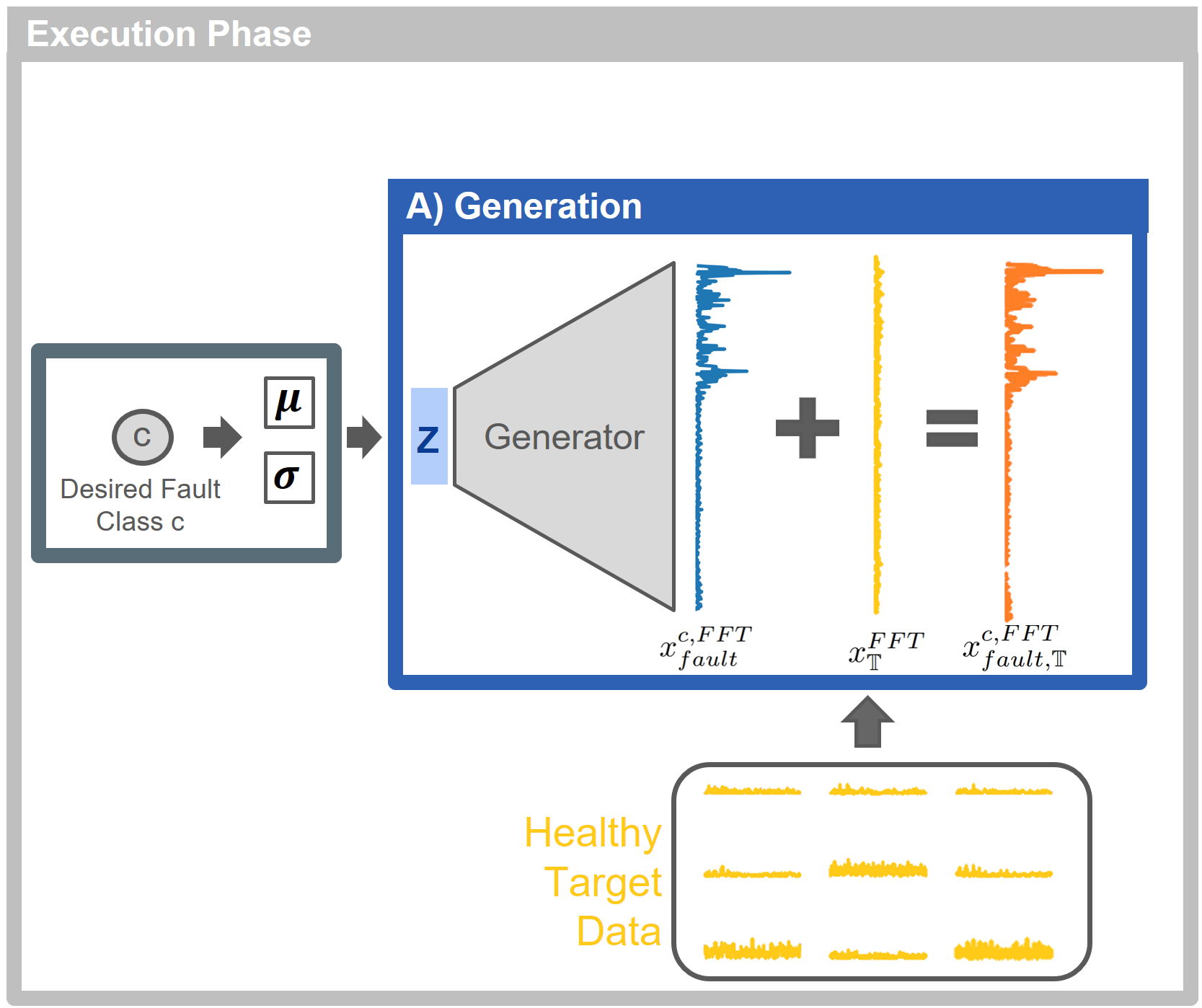}}
\caption{\label{fig:Step1} \textit{FaultSignatureGAN}: \textbf{Training Phase:} Training the A) generative model to generate domain independent fault characteristics while imposing B) plausibility with the discriminator in the source domain and C) semantic consistency with the classifier. \textbf{Execution Phase:} The generation of unseen target data.}
\end{figure}

\textit{FaultSignatureGAN} comprises three parts (A-C) as illustrated in \figref{fig:Step1}: (A) The first part ensures that generated fault signatures are easily transferable to a specific domain; (B) the second part ensures that the transformed fault signatures  represent plausible domain data;  and (C) the last part ensures that the transformed fault signatures are representative of the desired fault classes.  
Part \textbf{(A)} of the framework is tackled by a generative network that generates domain-independent fault signatures from distinct classes in the Fourier domain. These fault signatures are then transferred to a specific domain by adding them to randomly sampled data from the domain's healthy class. 
The ability of the generated data to represent true domain fault data is imposed in  part \textbf{(B)} by an adversarial discriminator. The semantic plausibility of the generated data to represent a desired fault class (as sampled from the sampling module) is tackled by a cooperative classifier in part \textbf{(C)}. 
The different parts of the framework are detailed below. 

\textbf{The underlying hypothesis: } 
The proposed approach is based on the hypothesis that the Fourier spectrum of fault data can be expressed as the sum of 1)  domain-specific components (the spectrum of a signal from normal operation) and 2) of  fault-specific components representing the specific fault characteristics. Further, we assume that the latter (spectrum of a signal representing the specific faulty condition) corresponds to a general domain-independent fault signature that is adjusted to new domains simply by linear scaling. In other words, this hypothesis allows us to express Fourier coefficients \cite{cooley1965algorithm}  of the fault data of a certain class $c$  from a specific domain $\mathbb{X}$  ($x_{fault,\mathbb{X}}^{c,FFT}$) as a sum of (1) domain-specific characteristics that are represented by the healthy class data of that domain $x_{\mathbb{S}}^{FFT}$ and (2) the fault class specific characteristics that are domain-independent $x_{fault}^{c, FFT}$ and scaled by a factor $w$- see \eqref{eq:source}.

\begin{align}
x_{fault,\mathbb{X}}^{c, FFT} &= x_{\mathbb{X}}^{FFT} + w * x_{fault}^{c, FFT} \label{eq:source}
\end{align}

The linear scaling with $w$ is performed to account for the fact that the fault signature is affected by operational changes and, therefore, we alleviate the strong assumption that the  fault-specific variations of real fault data are independent from operating conditions. Between a source domain $\mathbb{S}$ and target domain $\mathbb{T}$, the weight factor $w$ is defined as in \eqref{eq:factor}.
\begin{align}
w & = E(\mathbb{T}_{h,\mathbb{S}}/\mathbb{T}_{h,\mathbb{T}})
\label{eq:factor}
\end{align}

\textbf{(Part A) The generative model $G_{\theta}$: } The final goal is to generate faults in a target domain that have not been observed before. However,  from the two components of the faulty signal in the target domain as given in \eqref{eq:source}, we only have access to the healthy data representing the domain-specific variations ($x_{\mathbb{T}}^{FFT}$)  in the target domain $\mathbb{T}$. Therefore, to generate unseen faults in the target domain $x_{fault,\mathbb{T}}^{c,FFT}$, we need to design a framework that enables the  generation of the domain-independent characteristics of a fault class $x_{fault}^{c, FFT}$. 
 We train the proposed architecture on the data from the source domain, where we do not only have access to the healthy data $x_{\mathbb{S}}^{FFT}$ but also to true fault data $x_{fault,\mathbb{S}}^{c, FFT}$. In the source domain $\mathbb{S}$, the scaling factor equals to $1$.
  Due to the variability in the healthy class, simply subtracting the individual healthy samples from faulty ones (the reverse operation) is not sufficient to retrieve a domain independent fault signature. Therefore, we propose a generative model.
 The generative model is trained such that its output (blue signal in \figref{fig:Step1}) can be transformed in a real source fault by adding it to a healthy source sample (according to \eqref{eq:source}). 
 Thus, the generated signal can be transformed to real domain faults with any of the samples from the healthy data distribution. 
 In this study, we train one generative model to generate all severity levels of one fault type. This process is depicted in \figref{fig:Step1}. 
 To ensure plausibility of the generated signals in the specific domain, the generator is trained to fool \textbf{a discriminator $D_w$} (see below). To ensure semantic or class consistency, we condition the generative model on  the desired fault class by simply sampling the distinct desired fault class from a categorical distribution. Each of the discrete values drawn from the categorical distribution corresponds to a specific fault class. The probability of each category is defined based on the class distribution in the training dataset $T_{f,\mathbb{S}}$, from which the fault signatures should be learned from. In other words, the probability of category $i$ is defined by \eqref{eq:cat}.  The value sampled from the  uniform distribution is then passed to two vectors ($\mu$ and $\sigma$ in \figref{fig:Step1}), that parameterize a Gaussian distribution (mean and deviation), from which we sample using the reparametrization trick \cite{kingma2013auto}.   The generative model is updated based on the consistency of the desired class with the \textbf{the classifier's} prediction (see below).
 
 \begin{align}
p_i =  \frac{|\{(x_j, y_j) | ((x_j, y_j) \in T_{f,\mathbb{S}}) \& (y_j = i) \}|}{|T_{f,\mathbb{S}}|}
\label{eq:cat}
\end{align}
 To enable a better distinction, we will refer to the  signal representing the domain independent fault characteristics (depicted in blue in \figref{fig:Step1}) as the \textit{generated fault signature}, and to the  signal representing  the domain-specific fault data (depicted in red or orange in \figref{fig:Step1}) as  the\textit{ generated data sample }throughout the paper. Further, in the following, we will consider the data always in the Fourier domain without emphasizing it specifically. 

\textbf{(Part B) The discriminator $D_w$:} We need to ensure that the generated data represents plausible domain data. 
% Recap
Our final goal is to generate unseen data from a target domain. 
However, this target fault data has not been  observed so far. Hence, we 
cannot ensure plausibility of the generated data in the target domain directly while training the generative model. Instead, we train the generator to generate plausible  fault samples in the source domain. 
Therefore, the discriminator is trained to discriminate between real fault data of the source domain and the generated synthetic source data. 
% Why Wasserstein?
We implement a Wasserstein GAN \cite{arjovsky2017wasserstein} that is optimized with gradient penalty \cite{gulrajani2017improved} since its training has proven to be more stable compared to other GAN implementations,  mitigating mode collapse.
The adversarial loss function is defined by \eqref{eq:WGAN}. 

\begin{equation}
L_D = \mathbb{E}_{\tilde{x}^c, x_{h,\mathbb{S}}}[D_w(\tilde{x}^c + x_{h,\mathbb{S}})] - \mathbb{E}_{x_{f,\mathbb{S}}^c}[D_w(x_{f,\mathbb{S}}^c)] + \lambda_{GP}\mathbb{E}_{\hat x\sim\mathbb{P}_{\hat x}}[(||\nabla_{\hat{x}}D_w(\hat{x})||_2 -1)^2],
\label{eq:WGAN}
\end{equation}
where $D_w$ is the discriminator model, $\tilde{x}$ is a generated fault signature, $x_{h,\mathbb{S}}$ is a healthy source sample, $x_{f,\mathbb{S}}$ is a faulty source sample and $\hat{x}$ is drawn from $\mathbb{P}_{\hat x}$, a newly defined data distribution used to impose the gradient penalty.
For more details on the calculation of the gradient penalty, the interested reader is referred to \cite{gulrajani2017improved}.

\textbf{(Part C) The classifier $C_{\gamma}$:}
A classifier is added to the framework to ensure semantic consistency of the generated data to a desired class. The classifier is optimized with the semi-hard triplet loss \cite{schroff2015facenet} on real source data. In \eqref{eq:triplet}, the corresponding loss function is shown, where $C_{\gamma}$ is the classifier network, $\alpha$ is a fixed margin, $x_a$ is the anchor sample, $x_p$ the positive sample and $x_n$ the negative one.

\begin{equation}
L_{C} = max(||C_{\gamma}(x_a) - C_{\gamma}(x_p)||^2 - ||C_{\gamma}(x_a) - C_{\gamma}(x_n)||^2 + \alpha, 0)
\label{eq:triplet}
\end{equation}

For updating the generative model $G_{\theta}$, the semi-hard triplet loss is calculated using only  synthetic data $(\tilde{x} + x_{h})$ as anchors and real fault data $x_{f,\mathbb{S}}$ as positive resp.\ negative samples. 
A pseudo algorithm of the \textbf{Training Phase} is shown in \algoref{alg:1}.

\subsection{The generation of unseen data in the execution phase} 
After training the generative model $G_{\theta}$ in the \textbf{Training Phase}, the generation of target faults in the \textbf{Execution Phase} is straight forward: First, we sample the input of the generative model from a categorical distribution, which determines the desired fault classes to generate. The number of generated samples per class can be chosen freely. This input is then passed to the generative model to generate the respective fault class signatures. The fault signature is then transferred into the target domain (instead of the source domain) by (1) linearly scaling the fault signature (with $w$) and (2) adding it to the healthy data of the target domain (yellow data in \textbf{Execution Phase} of \figref{fig:Step1}). The scaling of the fault signature is defined as the ratio between the mean signal of the healthy source data and the mean of the healthy target data per frequency component (as defined in \eqref{eq:factor}).  Hence, the unseen target data is generated as defined in  \eqref{eq:target}. 
\begin{align}
x_{fault,\mathbb{T}}^{c, FFT} & = x_{\mathbb{T}}^{FFT} + w * x_{fault}^{c, FFT} \label{eq:target}
\end{align}

\begin{algorithm}
 \caption{Training Phase of \textit{FaultSignatureGAN}}
 \begin{algorithmic}
 \Require
 {\par  $T_{\mathbb{S}}$ (Source Dataset); $\lambda_{GP}, \lambda_{D}, \lambda_{E}, \alpha$ (Loss Function Parameter); $n_{critic}$, $es$ (Early Stopping Criteria), $m$ (Batch Size)}
 \Ensure  $G_{\theta}$
  %\State first statement
 %\\ \textit{LOOP Process}
 \LineComment{\textbf{Prepare Dataset}}
 \State $T_{h, \mathbb{S}} = \{ (x,y) \in T_{\mathbb{S}}$ $ |$ y is healthy $\}$; $T_{f,\mathbb{S}}= \{ (x,y) \in T_{\mathbb{S}}$ $ |$ y is a fault class $\}$
 \State $Cat(T_{f,\mathbb{S}})$ Categorical Distribution of the classes in $T_{f,\mathbb{S}}$
  \While{$es$ == False}
  \For{t = 1,..,$n_{critic}$}
   \LineComment{\textbf{Sample data batches}}
  \State $\{z^{(i)}\}_{i=0}^{m} \sim Cat(T_{f,\mathbb{S}})$
  \State $\{(x_{f, \mathbb{S}}, y_{f, \mathbb{S}})\}_{i=0}^{m} \sim \mathbb{T}_{f, \mathbb{S}}$; $\{x_{h, \mathbb{S}}\}_{i=0}^{m}\sim \mathbb{T}_{h, \mathbb{S}}$, $\{(x_{ \mathbb{S}}, y_{ \mathbb{S}})\}_{i=0}^{m} \sim \mathbb{T}_{\mathbb{S}}$ 
  \State $\epsilon \sim U[0,1]$
  \LineComment{\textbf{Generate data}}
  \State $\tilde{x} \leftarrow G_\theta(z)$
  \State $\tilde{x}_{f} \leftarrow \tilde{x} + x_{h,\mathbb{S}}$ 
  \State $\hat{x} \leftarrow \epsilon x_{f,\mathbb{S}} + (1-\epsilon)\tilde{x}_{f}$ 
  \LineComment{\textbf{Update discriminator D}}
  \State $L_{D}^{i} \leftarrow D_w(\tilde{x_f}) - D_w(x_{f,\mathbb{S}}) + \lambda_{GP}((||\nabla_{\hat{x}}D_w(\hat{x})||_2 -1)^2)$
  \State $w \leftarrow Adam(\nabla_w 1/m \sum_{i=1}^{m} L_{D}^{i},w)$
  \EndFor
  \LineComment{\textbf{Update classifier C}}
  \State From $\{x_{h, \mathbb{S}}\}_{i=0}^{m}$ form triplets \cite{schroff2015facenet} according to label $x_{ \mathbb{S},a}$, $x_{ \mathbb{S},p}$ and $x_{\mathbb{S},n}$
  \State $L_{C}^{i} \leftarrow max(||C_{\gamma}(x_{ \mathbb{S},a}) - C_{\gamma}(x_{ \mathbb{S},p})||^2 - ||C_{\gamma}(x_{ \mathbb{S},a}) - C_{\gamma}(x_{ \mathbb{S},n})||^2 + \alpha, 0)$
  \State $ \gamma \leftarrow Adam(\nabla_\gamma 1/m \sum_{i=1}^{m} L_{C}^{i},\gamma)$
  \LineComment{\textbf{Update generator G}}
  \State $L_D^{i} \leftarrow -D_w(G_\theta(z))$
  \State $L_C^{i} \leftarrow max(||C(\tilde{x}_f) - C(x_{ \mathbb{S},p})||^2 - ||C(\tilde{x}_f) - C(x_{ \mathbb{S},n})||^2 + \alpha, 0)$
  \State $L_{G}^{i} = \lambda_D * L_D^{(i)} + \lambda_C * L_C^{(i)}$ \State $\theta \leftarrow Adam(\nabla_\theta 1/m \sum_{i=1}^{m} L_{G}^{i},\theta)$
  \EndWhile \\
 \Return $G_{\theta}$
 \end{algorithmic}
 \label{alg:1}
 \end{algorithm}

\subsection{Alternative approaches used for comparison}
\label{sec:comparison}
In this work, we address two DA settings with label space discrepancies: \textit{Partial} DA and \textit{Open-Partial} DA, with a particular focus on the extreme case where only one class is shared between two domains. While for \textit{Partial} DA, some approaches have been proposed, only few are suitable for this extreme scenario. The approaches proposed in those few publications are used for comparison for the \textit{Partial} DA experiments. Besides (1) the \textit{Baseline}, we compare the proposed methodology to the following approaches: (2) a unilateral feature alignment approach in its fully unsupervised implementation, referred to as \textit{Unilateral} (2a), and in the implementation where the label of the target's healthy class is used for alignment, referred to as \textit{Unilateral$^{*}$} (2b) \cite{wang2020missing}, (3) a balanced and uncertainty-aware adversarial approach \textit{BA3US} \cite{liang2020balanced} as well as two data generation approaches - (4) \textit{GenAlign} \cite{li2018cross} and (5) \textit{PixelDA} \cite{bousmalis2016domain}. In general, the approach proposed to \citet{li2020deep2} would also be applicable in the extreme case of \textit{Partial} DA. However the achieved performance as reported on a comparable experimental setting (domain shift $2 \longrightarrow 3$ on CWRU), is inferior (90.1\%) compared to the results achieved with the \textit{Unilateral} approach (98.97\%). Therefore, this method is not listed as an comparison method in this study. 

(1) For the \textit{Baseline} comparison, we consider a setup where the classifier is trained on real source domain data only.  It shows the minimal achievable performance if no adaptation is performed. 
(2) The adversarial feature alignment approach \textit{Unilateral} DA \cite{wang2020missing} is chosen as a comparison method as it has been evaluated in a \textit{Partial} DA setting before (as elaborated in \secref{sec:RW}). It aims to achieve the same goals but uses a different strategy (feature alignment vs.\ data generation in our proposed framework). While originally proposed as a completely unsupervised DA method, the authors also  conducted experiments on the extreme scenario (where only the healthy class is shared between the two domains). For these experiments, the healthy data label from the target domain was used for alignment \cite{wang2020missing}. We compare our method to both implementations and denote the completely unsupervised implementation as \textit{Unilateral} and the  one using the target's healthy label for alignment as $Unilateral^{*}$. 
(3) The adversarial approach \textit{BA3US} balances each batch of target data with randomly sampled source data. It, therefore, presents an interesting comparison method to the proposed \textit{FaultSignatureGAN}, where we balance the target domain with generated data that has been mapped to the target domain in an unsupervised manner. Additionally, BA3US addresses the challenge of  uncertain class predictions in the source domain being propagated to the target domain.  The authors proposed to   exploit an adaptive weighted complement entropy objective to
encourage incorrect classes to have uniform and low prediction scores.
Moreover, we compare our method to two data generation approaches. 
(4) First, \textit{GenAlign} is used for comparison \cite{li2018cross} (see \secref{sec:RW}),  where target data is generated by passing novel input to the generative model. This approach is used to challenge the hypothesis that generative models are limited in their extrapolation abilities and therefore, the novel target data generation should not rely on extrapolation abilities of the model (as we do in our work).  
(5) The second data generation method  used for comparison is the domain mapping method
\textit{PixelDA} \cite{bousmalis2017unsupervised}. While it was originally applied to \textit{ClosedSet} UDA settings for DA tasks in computer vision (CV), one of the experiments in the paper was conducted on novel target classes that have not been represented in the training dataset. This setup resembles the \textit{Partial} DA setting. While the results obtained in the paper were very convincing, when applied to the extreme case of partial DA on CM data where only one class is shared between the domains, the good performance could not be replicated.  
Therefore, we modified the approach to allow for a fair comparison. Instead of generating the healthy target data (as proposed in the paper), we train the model to generate source faults, conditioned on the source healthy data. This has the following advantages: 1) the available source data is exploited better and 2) the novel input data to the generative model is more similar to the data the model has seen before. Ultimately, this modified \textit{PixelDA} approach enables to generate unseen target data by conditioning the generative model on the target healthy data at inference time. Since this data has not been seen before, the generation requires extrapolation abilities.   

For the \textit{Open-Partial} domain experiments, however, there is no other suitable comparison method that is applicable to the same extreme case scenario as we consider here where only the healthy class is shared between the two domains. Therefore, only the \textit{Baseline} is used for comparison in these experiments.

\section{Case Studies}
\label{sec:CaseStudies}
The proposed approach is tested on two bearing datasets that have been commonly applied for DA tasks in fault diagnostics in different settings. Our proposed framework is evaluated on both datasets in \textit{Partial} and \textit{Open-Partial} DA experiments. Both datasets are adjusted to the problem formulation to the respective DA setup.

\subsection{CWRU}
\label{sec:CWRU}
The CWRU Dataset is a publicly available benchmark bearing dataset (bearing type SKF 6205) provided by the Case Western
Reserve University Bearing Data Center (CWRU dataset) \cite{smith2015rolling}. The data was collected on a test rig in laboratory conditions. It contains data recorded under four different loads (referred to as domain 0,1,2 and 3). The different load settings resulting in different rotational speeds are shown in \tabref{tab:OCcaseStudies}. Data under healthy and nine different faulty conditions is available: Three fault types - Ball, Inner Race and Outer Race - with three severity levels each. An overview of the fault types and severity levels is shown in \tabref{tab:classes}. The CWRU dataset has been extensively used to demonstrate \textit{ClosedSet} DA methods under different operating conditions as well as for \textit{Partial} DA setups \cite{wang2020missing, li2020deep2}.

\subsection{Paderborn}
The Paderborn dataset is a publicly available bearing dataset (bearing type SKF 6203) provided by the Chair of Design and Drive Technology from Paderborn University \cite{lessmeier2016condition}. It incorporates both artificially induced bearing faults and realistic damages caused by accelerated lifetime tests \cite{zhang2020deep} under different operating conditions (rotational speed, load torque and radial force) \cite{chen2018acdin}. In this study, we only consider real fault data and not the artificially induced one. The represented health conditions in the dataset are healthy conditions, Inner Race faults (three severity levels) as well as Outer Race faults (two severity levels). The different operating conditions are shown in \tabref{tab:OCcaseStudies} and the different classes in \tabref{tab:classes}. The data was also collected on a test rig under laboratory conditions and was also previously used in different DA studies \cite{pandhare2019convolutional, chen2018acdin, wang2020missing}. Previous publications  mainly focused on 3-class classification  of the different fault types \cite{wang2020missing} and suggested that the domain gaps in the Paderborn dataset are larger compared to the CWRU dataset. Further, previous publications typically neglected the domain 1, since the domain gap to the other domains is considerably large compared to the other domain gaps. In this research, we focus on the type and severity classification (6-class classification) and also aim to bridge large domain gaps. Contrary to previous works,  we, therefore, included domain 1 in our DA evaluation. 

Moreover, we use less data compared to previous publications, such as e.g. \cite{wang2020missing}, for our evaluation (only using the datasets K002-5; KA04, KA15-16 and KI16,18,21 whereas KA22, KA30, KI04 and KI14 have not been used in this study). This enables us to evaluate if we can extract transferable fault characteristics  from only limited fault data. 

\ifdefined\ARXIV
    \begin{table}[h]
    \centering
    \begin{tabular}{l|c|c|c|c|c}
     & \multicolumn{1}{c|}{CWRU} & \multicolumn{4}{c}{Paderborn}\\\specialrule{2.5pt}{1pt}{1pt}
    Domain & \shortstack{Rotational\\ Speed [rpm]} & \shortstack{Rotational\\ Speed [rpm]} & \shortstack{Load \\ Torque [Nm]} & \shortstack{Radial  \\ Force [N]} & \shortstack{Setting   \\ Name} \\\specialrule{2.5pt}{1pt}{1pt}
    0 & 1797 & 1500 & 0.7 & 1000 & N15\_M07\_F10 \\
    1 & 1772 & 900 & 0.7 & 1000 & N09\_M07\_F10\\
    2 & 1750 & 1500 & 0.1 & 1000 & N15\_M01\_F10\\
    3 & 1730 &  1500 & 0.7 & 400 & N15\_M07\_F04\\
    \end{tabular}
    \caption{\label{tab:OCcaseStudies} Operating conditions under which the two case studies (CWRU and Paderborn) are recorded. Each setting corresponds to one domain. }
    \end{table}

    \begin{table}[h]
    \centering
    \begin{tabular}{l|l|c|c|c|c|c|c|c|c|c|c}
    & & Healthy &\multicolumn{3}{c|}{\shortstack{Outer Race\\ (OR)}} & \multicolumn{3}{c|}{\shortstack{Inner Race \\(IR)}} & \multicolumn{3}{c}{Ball (B)}\\\specialrule{2.5pt}{1pt}{1pt}
    \multirow{2}{*}{CWRU} & Size & - & 7 & 14 & 21 &  7 & 14 & 21 & 7 & 14 & 21  \\\cline{2-12}
     & Class & 0 & 1 & 2 & 3 & 4 & 5 & 6 & 7 & 8 & 9  \\\specialrule{2.5pt}{1pt}{1pt}
    \multirow{2}{*}{Paderborn} & \shortstack{Extent of \\ Damage} & - & 1 & 2 & - & 1 & 2 & 3 & - & - & -\\\cline{2-12}
    & Class & 0 & 1 & 2 & - & 3 & 4 & 6 & - & - & -\\
    \end{tabular}
    \caption{\label{tab:classes} Health conditions represented in the the case studies (CWRU and Paderborn).}
    \end{table}

\fi

\section{Experimental Setup}
\label{sec:ExpDomain} 
To test if \textit{FaultSignatureGAN} is capable of generating unseen domain faults, \textit{Partial} and \textit{Open-Partial} DA experiments are conducted, whereby the different domains correspond to the different operating conditions in the case studies. The experimental setups are shown in \tabref{tab:setuppartial} (\textit{Partial}) and \tabref{tab:setupopenset} (\textit{Open-Partial}) on an exemplary domain shift from some domain X to some target domain Y ($X \longrightarrow Y$) \tabref{tab:OCcaseStudies}.

\begin{table}[h]
\scriptsize
\centering
\begin{tabular}{cccccc}
  \toprule
  \multicolumn{6}{c}{\textit{Partial}}\\
  \toprule
  Dataset & \multirow{2}{*}{\shortstack{Domain\\Shift}} &  \multirow{2}{*}{\shortstack{Source\\Domain}} & \multirow{2}{*}{\shortstack{Source\\Classes}} & \multirow{2}{*}{\shortstack{Target\\Domain}} & \multirow{2}{*}{\shortstack{Target Classes\\during Training}} \\
  \cr
  \midrule
  CWRU & X $\rightarrow$ Y & X & 0,1,2,3,4,5,6,7,8,9 & Y & 0 \\
  Paderborn & X $\rightarrow$ Y & X & 0,1,2,3,4,5 & Y & 0 \\
 \end{tabular}
 \caption{\label{tab:setuppartial} Experimental Setup for \textit{Partial} DA on an exemplary domain shift $X \longrightarrow Y$.}
\end{table}

\begin{table}[h]
\scriptsize
\centering
 \begin{tabular}{cccccc}
  \toprule
  \multicolumn{5}{c}{\textit{Open-Partial}}\\
  \toprule
     \multirow{2}{*}{\shortstack{\\Task}} &\multirow{2}{*}{\shortstack{Domain\\Shift}} &  \multirow{2}{*}{\shortstack{Source\\Domain}} & \multirow{2}{*}{\shortstack{Source\\Classes}} & \multirow{2}{*}{\shortstack{Target\\Domain}} & \multirow{2}{*}{\shortstack{Target Classes\\during Training}} \\
     \cr
  \midrule
   \multirow{3}{*}{\shortstack{Source (IR)\\ $\rightleftharpoons$ \\Target (OR)}} & \multirow{3}{*}{ X $\rightarrow$ Y }  & \multirow{3}{*}{X} & \multirow{3}{*}{0,3,4,5} & \multirow{3}{*}{Y} & \multirow{3}{*}{0,1,2} \\
     \cr
   \cr
   \midrule
   \multirow{3}{*}{\shortstack{Source (OR)\\ $\rightleftharpoons$ \\Target (IR)}} & \multirow{3}{*}{ X $\rightarrow$ Y } & \multirow{3}{*}{X} & \multirow{3}{*}{0,1,2} & \multirow{3}{*}{Y} & \multirow{3}{*}{0,3,4,5} \\
   \cr
   \cr
  \bottomrule
\end{tabular}
\caption{\label{tab:setupopenset} Experimental Setup for \textit{Open-Partial} DA on an exemplary domain shift $X \longrightarrow Y$ on the Paderborn dataset.}
\end{table}

The experiments are conducted as follows: First, a generative model is trained on data from one domain (as described in \secref{sec:Methodology} and depicted in \textbf{Training Phase} in \figref{fig:Step1}). In this work, we train one generative model to generate all severity levels of one fault type.
Second, the label space of the target domain is completed by generating synthetic target fault data as depicted in \textbf{Execution Phase} in \figref{fig:Step1} based on healthy target data. The number of generated data samples per class is chosen to match the mean number of samples per class in the source domain.
In the third step, a new training dataset is composed of the generated and real data from both domains and used to train a classification model.  The performance of the classifier is then evaluated on a test dataset composed of all unseen faults and 30\% of the class data, from which the conditions have been observed before.

\textbf{Hyperparameter Tuning:}\label{sec:HPResults} Data-driven solutions based on neural networks come with many hyperparameters to tune including  those of the network architecture (layer type, activation, kernel size, initialization etc.\ ). These choices strongly influence the performance of the final model including its generalizability to new data.  
There is no commonly accepted procedure for optimizing the hyperparameters for an unknown target domain \cite{bousmalis2017unsupervised}. Some works rely, therefore, on a target validation dataset \cite{bousmalis2017unsupervised} or validation tasks \cite{wang2020missing}. In many practical applications, especially in the context of safety-critical systems, where no target fault data is available, this is not possible. Hence, in this work, we do not make the assumption of having  target data available for hyperparameter tuning, since it is a strong limitation of applying existing DA methods to real PHM applications. 

For training the \textbf{generative model} (\textbf{Training Phase} in \figref{fig:Step1}), only criteria related to the source dataset are used:  In addition to optimizing the loss functions (see \secref{sec:Methodology}) on the source dataset, a stopping criterion is implemented. The training is stopped if an auxiliary classifier trained on the synthetic source data returns an accuracy of at least 98\%, evaluated on the real source data. Since this callback function is computationally expensive, it is only executed after each 50 epochs of training.

\begin{figure}[h]
\centering
\includegraphics[width=1.0\textwidth]{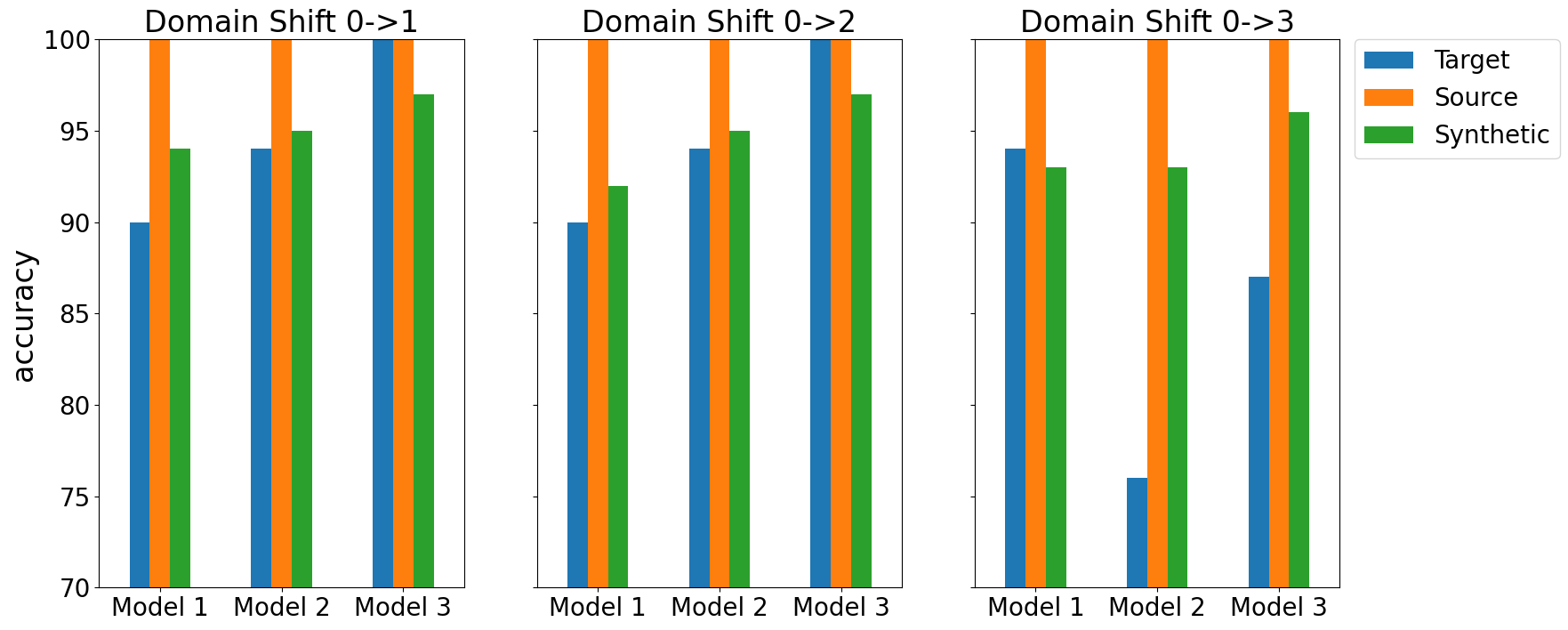}
\caption{\label{fig:LC}  Visualisation of the effect of hyperparameter tuning on the generalizability of different model architectures on different domain shifts: Model 1 as in \cite{wang2020missing}, Model 2 as in \cite{wang2020missing} but with the ReLu activation function, Model 3 as in \cite{wang2020missing} with the  kernel size of  12 (compared to 3 used in \cite{wang2020missing}). Three different datasets are used for evaluation: 1)source validation dataset (orange); 2) dataset with synthetic faults (green), and 3) the real target dataset (blue).}
\end{figure}

Further, the hyperparameters of the final \textbf{classification model} need to be tuned as well. In absence of real target fault data, we used synthetically generated data as a validation dataset. To showcase and evaluate the impact that hyperparameter settings have on the ability of the model to generalize to an unseen domain, we trained three different model architectures: Model (1) equals the one used in previous publications \cite{li2018cross, wang2020missing} , Model (2) equals Model (1) but  has the ReLu activation function and Model (3) equals Model (1) except that the kernel size is set to 12 (compared to 3 used in \cite{li2018cross}).  Exemplary, we only evaluate the  domain shifts from source domain 0 on the CWRU dataset for hyperparameter tuning. The final accuracies  on a source validation dataset, a synthetic fault dataset as well as on the true target test dataset of the three models are shown in  \figref{fig:LC}. 
The performance on  the target dataset varies considerably depending on the architecture used. For example, on domain shift $0 \rightarrow 2$, the final performance on the target dataset varies by 10$\%$ depending on the model used. This evaluation shows clearly that even small  changes in hyperparameters can impact the generalization ability strongly i.e.\ have a big effect on the performance in the target domain.
The source validation dataset does not provide a good indication which model to choose since the performance on the source dataset always results in $100\%$. The accuracy of the synthetically generated dataset does not correlate strongly with the target accuracy in all instances. However, it  gives a clear indication to choose Model 3 in all instances. This is also the best choice for the highest  accuracy on the target dataset in the first two domain shifts.
Only for domain shift $0 \rightarrow 3$ this is not an ideal choice. Although not ideal, we want to emphasize that the synthetic validation dataset provides information on which model to choose compared to the source validation dataset. On average, that information results in the best final model choice. Therefore, we conduct our experiments on Model (3) for the CWRU dataset. We train the classification model for 2000 epochs (since the source as well as the synthetic validation datasets suggest that no considerable change happens after 2000 epochs). To enable a better comparison, we use only one model architecture for all domain shifts per case study.  

Apart from using a synthetic validation dataset, we propose to use the following strategy for certain hyperparameters: 
(1) Applying a heavy regularization (since it leads to better generalization); (2) running the optimization for multiple epochs - more than indicated by the validation dataset. The latter choice is motivated by the findings of learning theory that hypothesize that there are two phases of deep learning: a fitting and a compression phase. It is indicated that the latter is responsible for the excellent generalization performance of deep networks \cite{shwartz2017opening}. Even though this hypothesis has been challenged recently
\cite{saxe2019information}, we still decided to set the number of optimization steps high.

The final model architectures being used and hyperparameter settings are elaborated in \secref{sec:appendix}.

\textbf{Label Availability:} 
Similarly to previous studies conducted on the extreme case of label discrepancy \cite{wang2020missing, li2020deep2}, we assume to know the label of the healthy data in the target domain for the \textit{Partial} and \textit{Open-Partial} DA experiments. Since healthy data is ubiquitous, this is considered to be a  realistic assumption. In addition, for the \textit{Open-Partial} setup, we assume that if at training data acquisition time fault classes have been observed, we also have the labels for the fault classes, both for source and target domains. This is a particularly realistic setup for fleets of complex systems with different ages for each of the units (each of the domains). Some units will have experienced one subset of fault types and other units will have experienced another subset of fault types. However, at testing time, we would like to be able to diagnose all of the fault types for all of the units.

\textbf{Data Pre-Processing:}
To enable a fair comparison, the datasets are pre-processed in the same way as in previous publications \cite{wang2020missing, li2018cross}. 
The CWRU datasets are first truncated (at 12000 timesteps) and divided into 200 sequences of 1024 points. After applying the Fast Fourier Transform \cite{cooley1965algorithm}, only the first 512 coefficients are used (excluding the first one).

The same process is applied to the Paderborn dataset. However, the data is not truncated and the 1024 long samples are sliced with a stride of 4096.

\section{Evaluation and Results}
\label{sec:Results}
\textit{Partial} (see \secref{sec:Partial}) as well as \textit{Open-Partial} DA experiments are conducted (see \secref{sec:OpenSet}) to test the ability of the generated data to bridge domain gaps in different DA settings. Last, to evaluate the physical plausibility of the generated data qualitatively, we visualise the generated data (see \secref{sec:visual}).

\subsection{Partial DA}
\label{sec:Partial}
\begin{figure}[h]
\centering
\subfloat[\centering  Real datasets. \label{fig:pad_pa_1}]{\includegraphics[ width=0.32\textwidth]{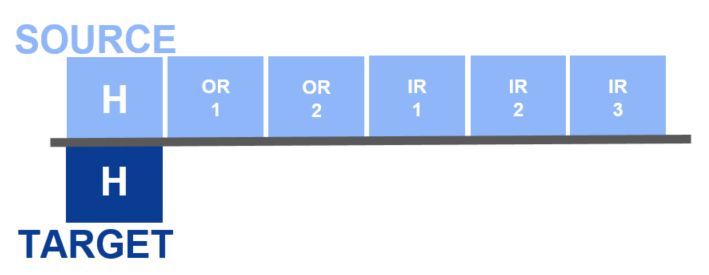}} 
\subfloat[\centering  Real and generated training datasets. \label{fig:pad_pa_2}]{\includegraphics[ width=0.32\textwidth]{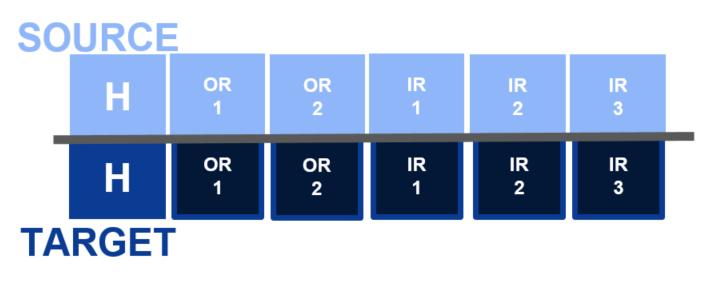}} 
\subfloat[\centering  Real test dataset. \label{fig:pad_pa_3}]{\includegraphics[ width=0.32\textwidth]{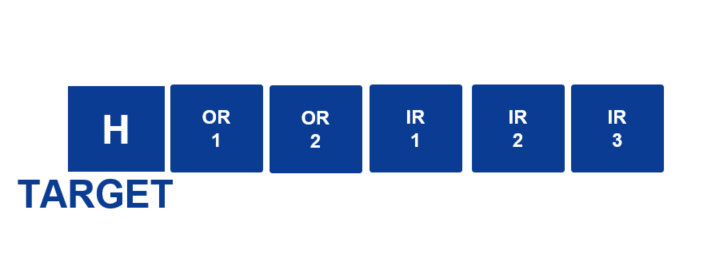}} 
\caption{\label{fig:Partial_Exp} Example of an extreme experimental \textit{Partial} DA Settings on the Paderborn case study. In the \figref{fig:pad_pa_1} The real datasets are shown where only the healthy class is shared between the source and the target dataset and the source dataset has five private classes. In \figref{fig:pad_pa_2} The training dataset is shown where the missing fault classes in the target domain are synthetically generated and in \figref{fig:pad_pa_3} the test dataset consisting of real target data is shown.}
\end{figure}

First, we conduct experiments in the extreme case of \textit{Partial} DA where only the healthy class is shared between the domains. The experimental setup is exemplified in \figref{fig:Partial_Exp} for the Paderborn dataset, where only the healthy data of the target domain is available. Therefore, we first generate the missing data in the target domain (darker blocks in  \figref{fig:Partial_Exp}) and evaluate on a target test dataset. 

\textbf{CWRU:}  We compare the results of the proposed methodology with the methods outlined in \secref{sec:comparison}. If available, the exact results from previous publications are shown. If not, we re-implement the methods while using exactly the same setup as in the original publications. Only for \textit{Ba3US}, a new architecture needed to be tuned as elaborated in \secref{sec:appendix} since this method has not yet been applied to any of the presented case studies. Since hyperparameter tuning without fault data  did not lead to satisfying results, we followed the protocol of \citet{wang2020missing} and used the domain shift experiment $0\longrightarrow3$ as a validation task.  For the source-only experiments, we report on the one hand the  previously reported results based on the originally proposed classifier architecture \cite{wang2020missing} with a kernel size of 3  (referred to as $Baseline$). On the other hand, we report the results of the source-only experiments conducted with the classifier architecture optimized based on the synthetic data as reported in \secref{sec:HPResults}. This architecture has a kernel size of 12 (referred to as $Baseline^{Syn}$).
The balanced accuracy of all experiments is shown in \tabref{tab:ExpDomainCWRU}. The overall performance of all approaches is high - even for the two source-only \textit{Baselines}, suggesting that the domain gaps are small in this dataset. This also leaves only limited room for improvement. The baseline model with optimized hyperparameters ($Baseline^{Syn}$), however, outperforms the existing $Baseline$ by $1.71\%$, showcasing that the generated data is beneficial for hyperparameter tuning. Moreover, adding the data generated by \textit{FaultSignatureGAN} to the training dataset, results in an additional improvement of $0.96\%$ - resulting in a total improvement of $2.67\%$ compared to the previously reported baseline method. This shows that the generated data is beneficial in order to bridge domain gaps. 

The two methods based on data generation used for comparison, (\textit{PixelDA$^{+}$} and \textit{GenAlign}), where the generative model is conditioned on novel input data (the real signal in \textit{PixelDA$^{+}$} and the target features in \textit{GenAlign}) perform worse than \textit{FaultSignatureGAN}. These results suggest that it is not beneficial  to condition the generative model on unseen input and rely on extrapolation abilities of the generative model as in \textit{GenAlign} and \textit{PixelDA}. Therefore, these approaches are not used as comparison methods in the following experiments on the Paderborn dataset. 
From the two unsupervised adversarial alignment approaches (\textit{Unilateral} and \textit{BA3US}), the \textit{Unilateral} approach performed consistently better on all domain shifts. 
%However, the unsupervised DA technique \textit{Unilateral} using the consistency loss improves the baseline performance by a small margin.  
\textit{Unilateral}$^{*}$, where the label of the healthy target data is used for alignment, results in the highest performance compared to all other approaches. On average, \textit{FaultSignatureGAN} performs within the same range of  \textit{Unilateral}$^{*}$.  In the following experiments on the Paderborn dataset, only \textit{Unilateral}$^{*}$ is used as a comparison method.  

\begin{table}[h]
\scriptsize
\centering
\begin{threeparttable}
\resizebox{\textwidth}{!}{%
\begin{tabular}{l|r|r|r|r|r|r|r|r} 
\shortstack{\tiny{Domain}\\\tiny{Shift}} & \multicolumn{1}{c|}{Baseline\tnote{1}} & \multicolumn{1}{c|}{Baseline$^{Syn}$}& \multicolumn{1}{c|}{Unilateral\tnote{2}}  &  \multicolumn{1}{c|}{Unilateral$^{*}$\tnote{1}}& \multicolumn{1}{c|}{BA3US}& \multicolumn{1}{c|}{\shortstack{Pixel \\DA$^{+}$}} & \multicolumn{1}{c|}{\shortstack{Gen\\Align \\\cite{li2018cross}}\tnote{1}}  & \multicolumn{1}{c}{\shortstack{\textit{Fault} \\ \textit{Signature} \\ \textit{GAN}}}  \\\hline

  0 $\rightarrow$ 1 & 93.49$\pm$1.75 & 99.49$\pm$0.06  & 97.04$\pm$0.86 & 98.08$\pm$0.16 & 91.07$\pm$3.98 & 92.45 & 97.81 & \textbf{99.87$\pm$0.07} \\
 
0 $\rightarrow$ 2 & 93.65$\pm$0.96 &  99.96$\pm$0.02 & 96.38$\pm$2.34 & \textbf{99.56$\pm$0.18} & 91.12$\pm$1.28 & 91.37  & 96.02 &  99.36$\pm$0.36 \\

0 $\rightarrow$ 3 & 91.02$\pm$0.02 & 90.27$\pm$0.69 &  94.14$\pm$0.56 & \textbf{98.22$\pm$0.65} &  96.33$\pm$1.71 &  86.01 & 94.24 &   94.50$\pm$1.10\\
\cline{1-9}

1 $\rightarrow$ 0 & 97.93$\pm$0.93 & 96.79$\pm$0.45 & 97.48$\pm$0.45 & 98.08$\pm$0.32 & 96.98$\pm$1.02 &\textbf{99.59} & 97.27 &    97.62$\pm$0.19  \\
1 $\rightarrow$ 2& \textbf{100.00$\pm$0.00} & \textbf{100.00$\pm$0.00} & \textbf{100.00$\pm$0.00}  & \textbf{100.00$\pm$0.00} & 99.95$\pm$0.04 & 99.38  &  96.32 &  99.95$\pm$0.00 \\

1 $\rightarrow$ 3 & 98.26$\pm$1.63 & \textbf{99.46$\pm$0.19} & 98.40$\pm$0.91  & 99.20$\pm$0.19& 99.23$\pm$0.66 & 93.81 & 94.59 & 99.35$\pm$0.11\\
 \cline{1-9}

2 $\rightarrow$ 0 & 91.63$\pm$1.63 & 96.15$\pm$0.15  & 90.13$\pm$3.66  & 96.43$\pm$0.43& 93.7$\pm$3.59  & 94.98 & 95.44 & \textbf{96.50$\pm$0.16}   \\

2 $\rightarrow$ 1 & 97.09$\pm$0.09 & 97.78$\pm$0.09 & 97.84$\pm$0.26  & 97.48$\pm$0.40& 95.58$\pm$1.93 & \textbf{98.97} & 96.55 & 97.06$\pm$0.09\\

2 $\rightarrow$ 3 & \textbf{99.78$\pm$0.17} & 99.63$\pm$0.12 & 99.71$\pm$0.10  & 98.97$\pm$0.21&  99.75$\pm$0.21 & 96.00 &  96.13 & 99.63$\pm$0.09 \\
 \cline{1-9}

3 $\rightarrow$ 0 & 87.96$\pm$0.18 & 88.58$\pm$0.19 & 86.50$\pm$4.56   & 94.85$\pm$2.16 & 86.75$\pm$0.21 & \textbf{96.65} & 92.82 & 92.81$\pm$0.92 \\

3 $\rightarrow$ 1 & 89.42$\pm$0.42 & 92.68$\pm$0.64 & 93.22$\pm$0.97  & \textbf{96.18$\pm$0.50} & 85.53$\pm$3.19 & 94.07 & 93.04 & 95.41$\pm$0.21\\
3 $\rightarrow$ 2  & 99.65$\pm$0.17 & 99.68$\pm$0.44 & 99.82$\pm$0.04  & 99.78$\pm$0.09 & 98.68$\pm$2.23 & 98.72 & 95.49 & \textbf{99.99$\pm$0.01}\\ \hline
 Mean & 94.99 & 96.70 & 95.88 &\textbf{98.07} & 94.56 & 95.17 & 95.49 &  97.66\\ \hline

\end{tabular}}
  \begin{tablenotes}
    \item[1] Results as reported in the original publication.
    \item[2] Models used as in original publication for reproducing results.
  \end{tablenotes}
\caption{\label{tab:ExpDomainCWRU} Extreme \textit{Partial} DA results on the CWRU dataset (10-class classification) under all domain shifts.}
\end{threeparttable}
\end{table}

\textbf{Paderborn:}
In this case study, we only use the best performing comparison DA method  \textit{Unilateral}$^{*}$. We neglect those that were not performing well on the CWRU case study. Moreover, contrary to other publications on the Paderborn case study that focus only on fault type classification (3-class classification),
we focus on the task of fault type and severity classification in this work (6-class classification).
Since the size of the domain gap differs considerably from \textit{domain 1 }to the other three domains (as indicated by the \textit{Baseline} results), we report the results for all DA tasks related to the domain 1 separately. Please note that in previous DA studies, the results on these DA tasks were never reported  \cite{wang2020missing}. Therefore, we report the results separately. In the upper part of  \tabref{tab:TwoExpDomainPaderborn}, only the results on \textit{domains 0,2} and \textit{3} are reported (smaller domain gaps) and in the lower part all results on DA shifts related to \textit{domain 1 }are reported (large domain gap).  

The DA approaches (\textit{Unilateral}$^{*}$ and \textit{FaultSignatureGAN}) outperform the \textit{Baseline} on all domain shift experiments - see \tabref{tab:TwoExpDomainPaderborn}. While the performance gain is comparable between the two approaches on \textit{domains 0,2} and \textit{3}, \textit{FaultSignatureGAN} results in a considerably better performance on \textit{domain 1}, where the domain gap is large (as indicated by the low \textit{Baseline} performance). In all settings, there is a substantial relative gain. On \textit{domains 0,2,} and \textit{3}, an average improvement of $3.82\%$ was achieved by \textit{FaultSignatureGAN} compared to the \textit{Baseline}. On \textit{domain 1}, the relative improvement is even $23.76\%$. The absolute performance differs between the different domain shift experiments: If \textit{domain 1} is the target domain,  the absolute performance of all approaches is still rather low ($<50\%$) despite the relatively high improvement. In the opposite direction, when\textit{ domain 1} is the source domain, higher absolute results were achieved (average performance of the three domain shift experiments with \textit{FaultSignatureGAN} is $79.72\%$). 
Although the domain gap should be the same in both directions (domain as source or target), this difference in the performance could potentially be explained if the fault data in \textit{domain 1 }shows more variability compared to the other domains.
This leads to better generalization on tasks from \textit{domain 1} to other domains. Especially in these instances, \textit{FaultSignatureGAN} shows a superior performance.

\begin{table}[h]
\scriptsize
\centering
\begin{tabular}{l|r|r|r}
 \shortstack{Domain\\Shift} & \multicolumn{1}{c|}{Baseline} & \multicolumn{1}{c|}{Unilateral$^{*}$} &  \multicolumn{1}{c|}{\textit{FaultSignatureGAN}}  \\\hline
%B & 0 & 0 & GAN+Triplet & 97 $\pm$ 1 \\
  0 $\rightarrow$ 2 & \textbf{99.78$\pm$0.06} & 99.92$\pm$0.02 &  99.76$\pm$0.24\\
   0 $\rightarrow$ 3 & 69.49$\pm$0.73 &  69.98$\pm$1.73 &  \textbf{72.08$\pm$1.24}\\
\cline{1-4}

  2 $\rightarrow$ 0 & 99.42$\pm$0.01  & \textbf{99.67$\pm$0.08} & 99.54$\pm$0.16   \\
  2 $\rightarrow$ 3 & 74.66$\pm$0.40  & \textbf{75.71$\pm$1.69} & 75.49$\pm$0.23 \\
 \cline{1-4}

  3 $\rightarrow$ 0 & 67.43$\pm$0.6 & 71.29$\pm$1.61  & \textbf{78.87$\pm$0.65} \\
  3 $\rightarrow$ 2  & 68.37$\pm$1.5 & \textbf{77.14$\pm$1.42} &  76.35$\pm$0.42\\ \specialrule{2.5pt}{1pt}{1pt}
Mean  &   79.86  &  82.29 &  \textbf{83.68}   \\ 
\specialrule{2.5pt}{1pt}{1pt}

 0 $\rightarrow$ 1 & 22.88$\pm$1.51 & 29.37$\pm$1.20 & \textbf{45.88$\pm$2.21} \\
 \cline{1-4}

 1 $\rightarrow$ 0 & 58.66$\pm$1.71 & 74.54$\pm$0.53 & \textbf{84.34$\pm$0.21} \\
   1 $\rightarrow$ 2&  63.28$\pm$1.77  & 75.16$\pm$3.56  &  \textbf{86.34$\pm$0.77} \\
   1 $\rightarrow$ 3 & 47.99$\pm$0.22  &61.87$\pm$1.77 &  \textbf{68.50$\pm$0.51}\\
 \cline{1-4}

  2 $\rightarrow$ 1 &   21.77$\pm$0.53  & 29.96$\pm$1.83 & \textbf{47.59$\pm$1.66}  \\
 \cline{1-4}

   3 $\rightarrow$ 1 & 22.77$\pm$1.31 & 26.47$\pm$0.24 & \textbf{45.30$\pm$0.61} \\ 
 \specialrule{2.5pt}{1pt}{1pt}
 Mean  & 39.23 &  49.56 & \textbf{62.99} \\
 \specialrule{2.5pt}{1pt}{1pt}
\end{tabular}
\caption{\label{tab:TwoExpDomainPaderborn} Extreme \textit{Partial} DA results on the Paderborn dataset (6-class classification). In the upper part all results with domain shifts including \textit{domains 0,2} and \textit{3 }are shown and in the lower part all domain shifts including \textit{domain 1}.}
\end{table}

\subsection{\textit{Open-Partial} Domain Experiments}
\label{sec:OpenSet}

\begin{figure}[h]
\centering
\subfloat[\centering  Real datasets. \label{fig:pad_os_1}]{\includegraphics[ width=0.32\textwidth]{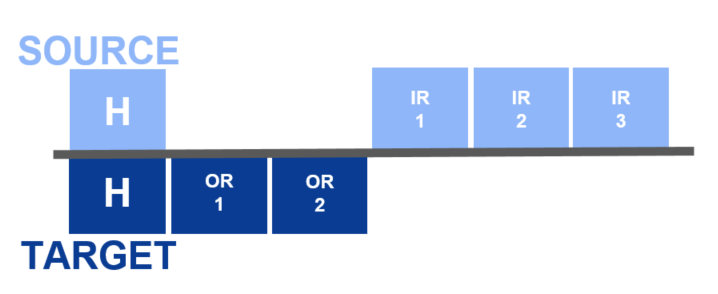}} 
\subfloat[\centering  Real and generated training datasets. \label{fig:pad_os_2}]{\includegraphics[ width=0.32\textwidth]{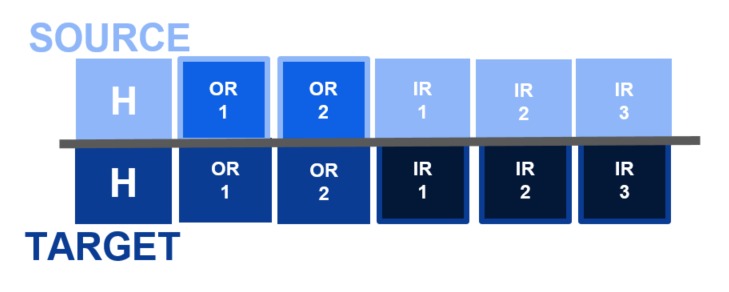}} 
\subfloat[\centering  Real test dataset. \label{fig:pad_os_3}]{\includegraphics[ width=0.32\textwidth]{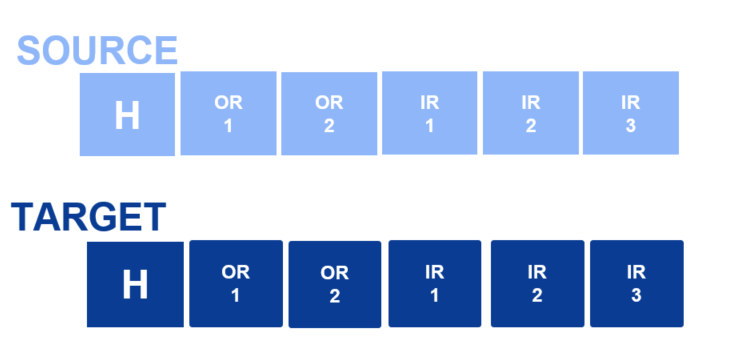}} 
\caption{\label{fig:Cross} Example of an extreme experimental \textit{Open-Partial} DA Settings on the Paderborn case study. In the \figref{fig:pad_pa_1} the real datasets are shown where only the healthy class is shared between the source and the target dataset and the source dataset has five private classes. In \figref{fig:pad_os_2} The training dataset is shown where the missing fault classes in the target domain are synthetically generated and in \figref{fig:pad_os_3}), the two test datasets are shown where one consists of the real source and the other of the real target data.}
\end{figure}

To showcase the versatility of our framework, we conduct \textit{Open-Partial} DA experiments in addition to the \textit{Partial} DA experiments (\secref{sec:Partial}). These experiments are only conducted on the Paderborn dataset since the domain gaps are larger compared to the CWRU dataset. The other DA methods used for comparison for the \textit{Partial} DA setup are not directly applicable for the \textit{Open-Partial} setup. Therefore, we only report the baseline results for  comparison.  \figref{fig:Cross} depicts an example for the the experimental setup: 
For the \textit{Open-Partial} DA experiments, we assume that in each of the two domains, one fault type occurred with different severities. The outer race fault with severity 1 and 2 (OR1 and OR2) occurred in the target domain, whereas the inner race fault (severity 1, 2 and 3; IR1, IR2 and IR3) occurred in the source domain. Hence, in a first step, two generative models are trained. The fault signature of the outer race fault is trained on the target data, the fault signature of the inner race fault on the source data (see \textbf{Training Phase} in  \secref{sec:Methodology}). In a second step, the missing fault data is generated: In the example of \figref{fig:Cross}, the outer race fault is generated for the source domain and the inner race fault classes with severity 1,2 and 3 for the target domain. This generated and real data composes the training dataset. 
Usually, only the performance on the target dataset is evaluated. However, in the experimental setup for \textit{Open-Partial} DA, there is missing data in each of the domains. Therefore, we evaluate the performance on two test datasets - the source $\mathbb{S}$ and the target $\mathbb{T}$. The test datasets comprise the real missing fault data as well as of a $30\%$ of known health conditions. The results on the 6-class classification task are reported in \tabref{tab:CrossPad}. The baseline was trained on all available real data from the source and target domain.   

The experiments show that the synthetically generated data enables to achieve a good classification performance on all settings ($>90\%$) excluding \textit{domain 1}, by far exceeding the performance of the \textit{Baseline} method of $83.37\%$ on average. On the DA task related to\textit{ domain 1}, the absolute performance of the classifier is considerably lower, however, it still results in a large relative improvement in all instances compared to the \textit{Baseline} method.

\begin{table}[h]
\scriptsize
\centering
\begin{tabular}{l|r|r|r|r|l|r|r|r|r|r|}
  & \shortstack{Domain\\Shift} & &\multicolumn{1}{c|}{Baseline} &  \multicolumn{1}{c|}{Proposed} &  & &\multicolumn{1}{c|}{Baseline} &  \multicolumn{1}{c|}{Proposed} \\\specialrule{2.5pt}{1pt}{1pt}

 \multirow{6}{*}{\shortstack{Source \\(IR)\\ $\rightleftharpoons$ \\Target \\(OR)}} & \multirow{2}{*}{0  $\rightleftharpoons$ 2} & $\mathbb{S}$ & 99.41$\pm$0.15 & \textbf{99.81$\pm$0.02} & \multirow{6}{*}{\shortstack{Source \\(OR)\\ $\rightleftharpoons$ \\Target \\(IR)}} &  $\mathbb{S}$ & 99.12$\pm$0.42 & \textbf{99.96$\pm$0.05}\\\cline{3-5}\cline{7-9}
 &  & $\mathbb{T}$ & 99.88$\pm$0.01   & \textbf{99.97$\pm$0.02} & & $\mathbb{T}$ & 99.53$\pm$0.14 & \textbf{99.85$\pm$0.10}\\
\cline{2-5}
  \cline{7-9}

 & \multirow{2}{*}{0 $\rightleftharpoons$ 3} & $\mathbb{S}$ & 73.65$\pm$0.13 &   \textbf{91.74$\pm$0.05} & & $\mathbb{S}$ & 78.99$\pm$0.69 &   \textbf{91.59$\pm$0.18}\\\cline{3-5}\cline{7-9}
 &  & $\mathbb{T}$ & 72.80$\pm$0.51 &    \textbf{95.33$\pm$0.01} & & $\mathbb{T}$ & 75.41$\pm$0.48 &  \textbf{96.18$\pm$0.75}\\
\cline{2-5}
  \cline{7-9}
 
    &\multirow{2}{*}{2 $\rightleftharpoons$ 3} & $\mathbb{S}$ & 75.30$\pm$1.04 &  \textbf{93.99$\pm$0.61} & & $\mathbb{S}$ & 78.10$\pm$1.17 &  \textbf{97.13$\pm$0.70}\\\cline{3-5}\cline{7-9}
 &  & $\mathbb{T}$ & 72.93$\pm$1.16 &    \textbf{94.82$\pm$0.23} & & $\mathbb{T}$ & 75.31$\pm$0.81&  \textbf{91.67$\pm$0.77}    \\
 \specialrule{2.5pt}{1pt}{1pt}
  
 \multirow{2}{*}{Mean} & & $\mathbb{S}$ & 82.79 & \textbf{95.18}  & & $\mathbb{S}$ & 85.40 & \textbf{96.22} \\\cline{3-5}\cline{7-9}
 &  & $\mathbb{T}$ & 81.87 & \textbf{96.71}    & & $\mathbb{T}$ & 83.42 & \textbf{95.90}    \\
  \specialrule{2.5pt}{1pt}{1pt}
  
 \multirow{6}{*}{\shortstack{Source \\(IR)\\ $\rightleftharpoons$ \\Target \\(OR)}}  &  \multirow{2}{*}{0 $\rightleftharpoons$ 1} & $\mathbb{S}$ & 56.91$\pm$1.44 & \textbf{66.55$\pm$0.61} &  \multirow{6}{*}{\shortstack{Source \\(OR)\\ $\rightleftharpoons$ \\Target \\(IR)}} & $\mathbb{S}$ &  51.53$\pm$1.43 & \textbf{53.90$\pm$1.33}\\\cline{3-5}\cline{7-9}
 &  & $\mathbb{T}$ & 65.53$\pm$0.09  &   \textbf{71.04$\pm$0.22} & & $\mathbb{T}$ & 69.06$\pm$1.28 &  \textbf{81.48$\pm$1.53}\\\cline{2-5}
  \cline{7-9}
   
   & \multirow{2}{*}{1 $\rightleftharpoons$ 2} & $\mathbb{S}$ & 53.62$\pm$0.23 & \textbf{56.21$\pm$1.18} &  & $\mathbb{S}$ & 51.93$\pm$1.21 &  \textbf{65.65$\pm$1.10}   \\\cline{3-5}\cline{7-9}
 &  & $\mathbb{T}$ & 69.71$\pm$0.16 &    \textbf{76.08$\pm$1.41} &  & $\mathbb{T}$ & 65.76$\pm$0.26 &  \textbf{67.54$\pm$0.34}\\
\cline{2-5}
  \cline{7-9}
  
  &\multirow{2}{*}{1 $\rightleftharpoons$ 3} &$\mathbb{S}$ & 51.83$\pm$1.66 &  \textbf{66.10$\pm$1.40} &  & $\mathbb{S}$ & 63.66$\pm$0.56  &  \textbf{67.64$\pm$0.15} \\\cline{3-5}\cline{7-9}
 &  & $\mathbb{T}$ & 66.50$\pm$0.02 &  \textbf{74.75$\pm$0.65} & & $\mathbb{T}$ & 65.25$\pm$0.06  &   \textbf{71.14$\pm$0.62}  \\
 \specialrule{2.5pt}{1pt}{1pt} 
  
 \multirow{2}{*}{Mean} & &$\mathbb{S}$ & 54.12 & \textbf{62.95}  & &$\mathbb{S}$ & 55.71 & \textbf{62.39} \\\cline{3-5}\cline{7-9}
 &  & $\mathbb{T}$ & 67.24 &  \textbf{73.96}  & &$\mathbb{T}$ & 66.69 & \textbf{73.39}    \\
  \specialrule{2.5pt}{1pt}{1pt}
 
\end{tabular}
\caption{\label{tab:CrossPad} \textit{Open-Partial} DA results on the Paderborn dataset (6-class classification). In the upper part all results with domain shifts including domains 0,2 and 3 are shown and in the lower part all domain shifts including domain 1. As shown in \figref{fig:Cross}, the trained classification model is evaluated on two datasets: The source test dataset ($\mathbb{S}$) and the target test dataset ($\mathbb{T}$).}
\end{table}

\subsection{Qualitative evaluation}
\label{sec:visual}
\begin{figure}[h]
\centering
\subfloat[\centering Mean Fault Signal OR Severity 1 of the synthetic data (blue line), the true target fault class data (orange line) and the true source fault class data (green line).\label{fig:MeanSignal}]{\includegraphics[width=0.75\textwidth]{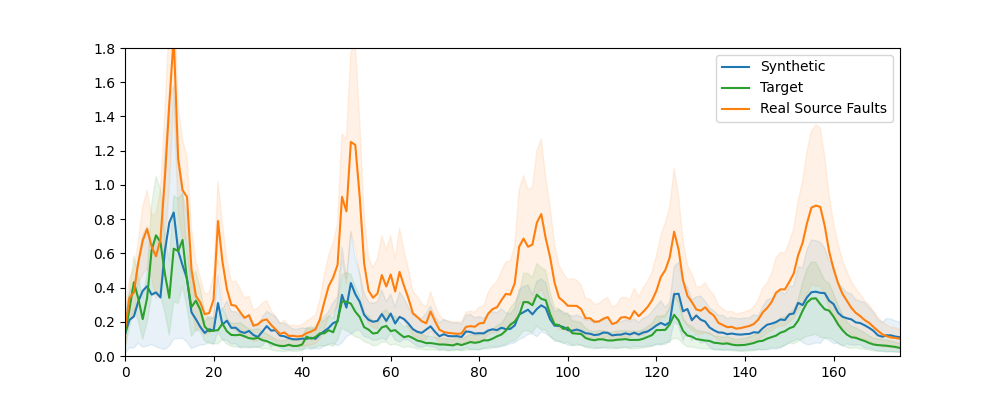}}\\
\subfloat[\centering Mean Absolute Residual Signal OR Severity 1 of the Target to Synthetic  (red line) and Target to Source (purple line).\label{fig:MeanError}]{\includegraphics[width=0.75\textwidth]{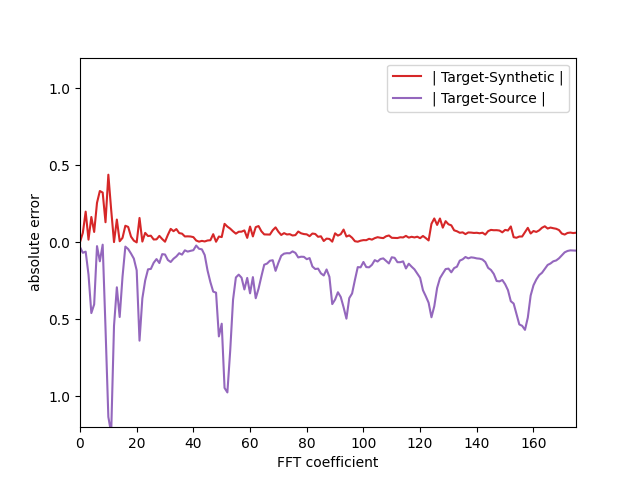}}
\caption{\label{fig:VisPA3} Paderborn data visualization of the OR severity 1 fault comparing real fault data with generated fault data.}
\end{figure}

To evaluate the physical plausibility qualitatively, we visualise the mean of the generated signals (blue line in \figref{fig:MeanSignal}), of the true faults in the target domain (orange line in \figref{fig:MeanSignal}) and of the true faults in the source domain (green line in \figref{fig:MeanSignal}). A batch of 1000 data samples was used for the illustration including its standard deviation.  Exemplary, we chose the \textit{fault type OR} and \textit{severity 1} on the  domain shift $0 \rightarrow 1$ on the Paderborn dataset. To better visualize the differences, the residual of the generated target to the true target mean signal is visualized as well as the residual of the true source to the true target (which can be considered as the baseline) - see \figref{fig:MeanError}. The proposed framework appears to generate the true target data considerably well. It performs substantially better compared to the baseline (just using the source faults without any adaptation for the target domain). Especially in the higher frequency range, it represents the true target faults noticeably better than the true source faults. 

\section{Discussion}
\label{sec:discussion}
The experiments performed in this research demonstrate the validity of the proposed framework \textit{FaultSignatureGAN} for DA with different types of extreme label discrepancies, where only the healthy class is shared between the domains. The results show that \textit{FaultSignatureGAN} performs similarly well compared to other approaches given small domain gaps and outperforms all comparison methods under large domain gaps in extreme \textit{Partial} DA experiments. Further, \textit{FaultSignatureGAN} is not limited to only one label space discrepancy setting (if labels are available) as demonstrated by the conducted \textit{Open-Partial} experiments, where the proposed method substantially outperforms the comparison method. The obtained results open interesting points for discussion. 

\textbf{Suitability of \textit{FaultSignatureGAN} for DA with label space discrepancies:} Given small domain gaps, \textit{FaultSignatureGAN} outperforms most of the comparison methods, especially when comparing the results to other generative approaches (\textit{PixelDA} and \textit{GenFeat}). This particularly supports the assumption that the unsupervised generation of unseen data should not rely on extrapolation abilities of the generative model (as it does for the comparison methods). Instead, our approach enables the generation of unseen faults building on the hypothesis that domain-specific fault data can be disentangled in domain-specific characteristics and class specific ones and therefore, requires no extrapolation ability of the generative model. The experiments show that the data generated by \textit{FaultSignatureGAN} results in a better diagnostic performance on the conducted experiments compared to the other generative approaches \textit{GenAlign} and \textit{PixelDA}, indicating that it resembles the true target data better. The balanced and uncertainty-aware method \textit{BA3US} outperforms \textit{FaultSignatureGAN} solely on domain shift $0\longrightarrow3$ (by 1.83\%). This is the domain shift where the hyperparameters of \textit{BA3US} were tuned on while \textit{FaultsignatureGAN} did not rely on this information because it is typically not available at training time, especially given safety-critical applications. However, the improvement that \textit{BA3US} provided on that one domain shift could not be translated to other domain shifts. This, once more, showcases the importance of hyperparameter tuning for DA methods based on feature alignment and in particular the importance of the access to fault data for hyperparameter tuning. 
Only the feature alignment approach \textit{Unilateral$^{*}$} provides a similar performance as \textit{FaultSignatureGAN} under small domain gaps in \textit{Partial} DA settings (see all results on CWRU in \secref{sec:Partial} and on Paderborn with domains 0,2 and 3).
This is not surprising since synthetic data generation is never perfect. Therefore, when the domain gap is small, the source data represents the target data already quite well  and one would expect little benefits in generating synthetic target specific data.
On large domain gaps (those including domain 1 on the Paderborn dataset), however, the performance of the feature alignment method \textit{Unilateral} drops. These are the scenarios where the proposed generative approach \textit{FaultSignatureGAN} outperforms other approaches (see \secref{sec:Partial} and \secref{sec:OpenSet}). 
Therefore, if the size of the domain gap is unknown, \textit{FaultSignatureGAN} is the best option to choose since it provides performs a comparable performance under small domain gaps but a considerably better performance under large domain gaps.

\textbf{Versatility of \textit{FaultSignatureGAN}:} Many different scenarios of label discrepancies are possible in real operations as exemplified in \secref{sec:Introduction}. Having one versatile method that can be applied in multiple of these scenarios is, therefore, utterly important for practical applications. The versatility of the proposed approach is demonstrated by applying it successfully to  DA experiments with different types of label discrepancies (\textit{Partial} and \textit{Open-Partial}) (if the respective labels are known in the source domain), where it consistently outperforms other comparison methods under large domain gaps. We consider the versatility of the proposed approach as a one of the key benefits for practical PHM applications.

\textbf{Plausibility of Unsupervised Data Generation and Validity of the Underlying Hypothesis:} The generation of unseen target data requires unsupervised data generation. The plausibility of the target data is not imposed while training the generative model and therefore, cannot be guaranteed. Therefore, it is required to evaluate how realistic  the target data generated by \textit{FaultSignatureGAN} is; to which extent it can be used as a surrogate of real target data. The data visualization (see \secref{sec:visual}) shows that the generated data represents real target data well. In particular, it represents the target fault data substantially better compared to the source data. 
This finding is also supported by the findings  in the DA experiments (both \textit{Partial} and  \textit{Open-Partial}) where \textit{FaultSignatureGAN} consistently outperforms the \textbf{Baseline} method. This supports the validity of the underlying hypothesis. We can draw the following conclusions: (1) \eqref{eq:source} serves as a good approximation of real fault data, (2) the disentanglement of domain-specific and fault-specific characteristics was successful and (3) that domain-invariant fault signatures can be extracted by \textit{FaultSignatureGAN} given only one source domain. However, our assumption about the structure of domain-specific and fault-specific components composing real fault data as defined in  \eqref{eq:source} could be extended and further refined in future work, in particular, the assumption  that the operating conditions impact the fault-specific components linearly.  Moreover, the DA experiments show that the generative process succeeds in preserving the semantic meaning of the generated data. If this would not be the case, the generated data would introduce label noise to the training data and, quite likely, result in a performance drop in the target domain.

\textbf{Synthetic Data for Hyperparameter Tuning:} We found that the synthetically generated  fault data can be used for hyperparameter tuning; potentially even for any DA model. This is an additional key advantage of the proposed data generative approach for practical applications compared to feature alignment methods for DA with label space discrepancies. Since the classification performance of the \textbf{Baseline} method in the target domain  is highly dependent on the chosen classifier architecture (see evaluation in \secref{sec:HPResults}), synthetic fault data can support in selecting the optimal architecture without relying on labels in the target domain what are usually not available in real applications, especially safety-critical ones. The discrepancy of the performance is also observed in the literature, where the hyperparameter search resulted in different architectures and hence, different \textbf{Baseline} results on exact same tasks but with different hyperparameters of  classifier architectures. For example, \citet{wang2020missing} reported a mean performance $99.78\%$ on the domain shift $2\rightarrow3$, whereas \cite{li2020deep} reported a baseline performance of $92.2\%$ using a different classifier architecture. The impact of the architecture choice or hyperparameter setting on the classification performance in the target domain is even larger than the relative improvements reported by other DA approaches. This emphasizes the importance of hyperparameter tuning including the choice of the network architecture for the task of DA. 
In absence of fault target data, there is no possibility to tune these hyperparameters with respect to the classification task in the target domain.
If, however, synthetic data is available that represents the real target data well, the data can be used for validation. Herein lays one major benefit of the proposed data generative approach \textit{FaultSignatureGAN}.  Although a proof of optimality is impossible (as the real target data has not been observed), the synthetic data  provides a better indication of which hyperparameters to choose compared to the hyperparameter choice based on the source dataset performance or even a random choice.

\textbf{Decreasing Data Acquisition Time:} 
In practice, a short data acquisition phase is essential to enable to start monitoring the condition of a new asset within a short period of time. However, faults are extremely rare in complex (safety-critical) systems. This lack of real fault data is a major limitation to applying data-driven solutions for fault diagnostics. \textit{FaultSignatureGAN} allows to transfer fault patterns to a new target domain. Once a fault occurred in one domain providing sufficient fault data to train a generative model, the fault signature can be learned, which then can be used to generate new fault data for any newly emerging domain. This ultimately can speed up the data collection process significantly, enabling the application of data-driven solutions within a shorter time span.

\section{Conclusion}
\label{sec:Conclusion}
In this research, we proposed the  \textit{FaultSignatureGAN} framework for controlled generation of unseen faults in the target domain. The resulting generated fault data is (1) specific to a desired domain and (2) specific to a certain fault type and the severity level of the fault in that domain. Therefore, \textit{FaultSignatureGAN} enables to start monitoring the condition of new assets without any faults observed in the target domain since plausible faulty data can be generated for all future target domains. While we considered different operating conditions as domains in this research, the proposed framework is also applicable to generate synthetic faults in new units of a fleet.

We demonstrated the potential of the \textit{FaultSignatureGAN} to complement partial label spaces in different DA experiments - \textit{Partial} as well as \textit{Open-Partial} DA settings. The results show that the generated data represents true faults in the target domain considerably better than the source fault data, leading to an improved classification performance on the target domain. Our proposed method excels particularly on large domain gaps. 
\textit{FaultSignatureGAN}  also enabled  hyperparameter tuning for unseen target domain which can be applied in combination with any other DA approach. Without any access to target faults, tuning existing methods optimally is not possible. This demonstrates one of the benefits of plausible data generation in the evaluated tasks.   

For future work, an additional step integrating real but unlabeled target data in addition to the synthetically generated data is an interesting direction to explore. Additional unsupervised or semi-supervised DA approaches could be employed to bridge the synthetic to real gap. Furthermore, the transferability of the generated fault signatures between different bearing types is of high interest for future research. One further direction of future research would be to investigate the source data demand for \textit{FaultSignatureGAN}, evaluating how many samples and how diverse they need to be in order to train a representative generative model. On a bigger scale, the integration of novel or evolving fault detection (those that have not been observed neither in the source nor in the target domain) in addition to the performed fault classification would be of a significant practical relevance.

\appendix
\section{Models}
\label{sec:appendix}
Unless stated otherwise, the following model architectures were used: 
\begin{description}
\item[Generation Model:] The first layer of the generation model is a single neuron. The activation of this neuron is sampled from a categorical distribution corresponding to the number of fault classes (fault type severities). The second fully connected layer is the sampling layer (mean and variance), containing three units each, activated by LeakyReLu ($\alpha$=0.001).

The following fully connected layers successively increase the dimensionalty to the desired final output shape. Each layer is activated by LeakyReLu ($\alpha$=0.001), using no bias and followed by a BatchNormalization (BN) layer.

Three 1D convolutional layers follow, each layer is activated by LeakyReLu ($\alpha$=0.001), and followed by a BatchNormalization (BN) layer.

At last the generated signal is added to a randomly drawn data point from the base dataset.

The Adam optimizer used with a learning rate of 0.0001, $beta_1=0.5$ and $beta_2=0.999$.

\item[Triplet Encoder Model:] The triplet encoder model consists of 6 fully connected layers, each activated with Leaky ReLu (alpha=0.1) and followed by a dropout layer (rate=0.4). The final layer is 4 dimensional and is L2 normalized. 

The Adam optimizer used with a learning rate of 0.0001, $beta_1=0.5$ and $beta_2=0.999$.

\item[Discriminator Model:] The discriminator model consists of six fully connected layers, each activated with Leaky ReLu (alpha=0.1) and followed by a dropout layer (rate=0.1). The final layer is 1-dimensional.
The Adam optimizer used with a learning rate of 0.0001, $beta_1=0.5$ and $beta_2=0.999$.
\item[Classification Model for Early Stopping:] It consists of 4 1-D convolutions layers (8 filters in each layer and kernel size is 3), each activated with Leaky ReLu (alpha=0.1) and followed by a dropout layer (rate=0.1). Followed by a flattening layer and a fully connected layer with the appropriate number of unots according to the number of classes in the dataset. 

The Adam optimizer used with default parameters.

\item[Classification Model for Evaluation:] The classification model for evaluation is inspired by \citet{wang2020missing} . It consists of three 1D convolutional layers (10 filters in each layer, activated by ReLu) and dropout layers (0.4). 
The Adam optimizer used with default parameters.

For training the CWRU classification models a batch size of 64 is chosen, for Paderborn 2000.

\item[Model Architecture for the comparison method \textit{BA3US}:] The comparison method \textit{BA3US} has not yet been applied to any timeseries data. Without using target fault data, the methodology could not be tuned to give satisfying results. Therefore, we followed the proceedure of \citet{wang2020missing} and tuned  \textit{BA3US} on a validation task 0$\longrightarrow$3.  We started using the exact same generator, discriminator and classifier architecture as well as optimizer setting as proposed by \citet{wang2020missing}. All hyperparameters (model architecture and weighting of the different loss terms) are then consecutively optimized on the validation task. Ultimately, the following architecture was used: The \textbf{feature extractor} consists of a 3 layer 1D-convolutional layer with kernel size 3 and 10 filters per layer. Each layer is batch normalized and activated by the sigmoid function, followed by a dropout layer (rate$=0.5$). Last, based on the flattened activations, a fully connected layer is added with 256 units. The \textbf{classifier} model consists of two fully connected layers. The first with 256 units is activated with the ReLU activation function and followed by a dropout layer (rate 0.5). The second contains 10 units (corresponding to the number of classes) and is activated by the softmax function. The \textbf{discriminator} contains three fully connected layer, each with 256 ReLU activated units. Only the last layer contains only one unit and is Sigmoid activated. The model is optimized on batches of 64 samples in the target and the source domain using the StochasticGradientDescent algorithm with a learning rate of 0.005. The initial ratio of augmented source samples is set to 1.0 ($\rho_{0}=1$), the test interval $N_u$ is set to 50. The loss conditional entropy loss is weighted with a factor of $10^{-3}$ and the transfer loss with a factor of $10^{-1}$. The weighted complement entropy loss is not considered since it did not lead to satidfying results ($w=0$).   

\end{description}

\section*{Acknowledgement}
This research resulted from the "Integrated intelligent railway wheel condition prediction" (INTERACT) project, supported by the ETH Mobility Initiative.

\printbibliography
\end{document}